\documentclass[journal,twoside]{IEEEtran}

\ifCLASSINFOpdf
  \usepackage[pdftex]{graphicx}
\else
\fi
\usepackage[cmex10]{amsmath}
\usepackage{multirow}
\usepackage{dsfont}
\usepackage{caption}
\def\best{\textbf}
\def\second{}

\def\M{{\mathbf M}}
\def\Y{{\mathbf Y}}

\def\J{{\mathbf J}}

\def\X{{\mathbf X}}
\def\Z{{\mathbf Z}}
\def\one{{\mathds 1}}
\def\seita{{\boldsymbol \theta}}

\begin{document}
%
\title{DeepSaliency: Multi-Task Deep Neural Network Model
for Salient Object Detection}
%
%
%

\author{Xi~Li,
        Liming~Zhao,
        Lina~Wei,
        Ming-Hsuan~Yang,
        Fei~Wu,
        Yueting~Zhuang,
        Haibin~Ling,
        and~Jingdong~Wang
\thanks{%
Manuscript received October 19, 2015; revised April 15, 2016 and June 1, 2016; accepted June 6, 2016. 
This work was supported in part by the National Natural Science Foundation of China under Grant 61472353 and Grant U1509206, 
in part by the National Basic Research Program of China under Grant 2012CB316400 and Grant 2015CB352302, 
in part by the Fundamental Research Funds for the Central Universities. 
The work of M.-H. Yang was supported in part by the NSF CAREER Grant 1149783 and the NSF IIS Grant 1152576. 
The work of H. Ling was supported in part by the National Natural Science Foundation of China under Grant 61528204 and the NSF IIS Grants 1218156 and 1350521. 
The associate editor coordinating the review of this manuscript and approving it for publication was Dr. Nikolaos Boulgouris. 
\textit{(Corresponding author: Liming Zhao.)}%
}%
\thanks{X. Li, L. Zhao, L. Wei, F. Wu and Y. Zhuang are with College of Computer Science, Zhejiang University, Hangzhou, China (email: $\{$xilizju,zhaoliming,linawzju$\}$@zju.edu.cn,  $\{$wufei,yzhuang$\}$@cs.zju.edu.cn).}%
\thanks{M.-H. Yang is with Electrical Engineering and Computer Science at University of California, Merced, United States (email: mhyang@ucmerced.edu).}%
\thanks{H. Ling is with the Department of Computer and Information Sciences, Temple University, Philadelphia, USA  (email: hbling@temple.edu).}%
\thanks{J. Wang is with Visual Computing Group, Microsoft Research Asia, China (email: jingdw@microsoft.com).}%
\thanks{Color versions of one or more of the figures in this paper are available online at http://ieeexplore.ieee.org.}%
}
\markboth{IEEE Transactions on Image Processing,~Vol.~XX, No.~X,~201X}%
{Li \MakeLowercase{\textit{et al.}}: DeepSaliency: Multi-Task Deep Neural Network Model for Salient Object Detection}
\maketitle

\begin{abstract}
A key problem in salient object detection is how to effectively model the
semantic properties of salient objects
in a data-driven manner.
In this paper, we propose a multi-task deep
saliency model based on a fully convolutional
neural network (FCNN) with global input (whole raw images)
and global output (whole saliency maps).
In principle, the proposed saliency model takes a data-driven strategy
for encoding the underlying saliency prior information, and then
sets up
a multi-task learning scheme for
exploring
the intrinsic correlations
between saliency detection
and semantic image segmentation.
Through collaborative feature learning from such two correlated tasks,
the shared fully convolutional layers produce effective features for
object perception.
Moreover, it is capable of capturing
the semantic information on salient objects across different levels
using the fully convolutional layers, which
investigate
the feature-sharing properties of salient object detection with
a great reduction of
feature redundancy.
Finally, we present a graph Laplacian regularized nonlinear
regression model for  saliency refinement.
Experimental results demonstrate the effectiveness of our approach
in comparison with the state-of-the-art approaches.
\end{abstract}



%

\section{Introduction}
\IEEEPARstart{A}{s} an important and challenging problem in computer vision, salient object detection~\cite{Contrast09, SVO, MC, GC, XL13, Sun_TIP2015,Zou_TIP2015,Borji_TIP2015}
aims to automatically discover and locate the visually interesting
regions that are consistent with human perception.
It has a wide range of applications
such as object tracking and recognition, image compression, image and
video retrieval, photo collage, video event detection, and so on.
The focus of salient object detection
is on designing various computational models to measure image
saliency, which is useful for segmentation.

From the perspective of human perception, the semantic properties of
an image are typically characterized by the objects and their
contexts within the same scene.
In principle, each object can be represented on three different levels
(i.e., low-level, mid-level, and high-level).
The low-level visual cues~\cite{IT98,Frintrop2015} are concerned
with primitive image features such as color, edge, and texture.
The mid-level features~\cite{CSprior, GS} typically
correspond to the object information on contour, shape, and spatial
context.
In comparison, the high-level information is associated with the object
recognition, segmentation as well as the intrinsic semantic
interactions among objects along with background~\cite{LowRank, SalDeepFeature,Li2015,Gong2015}.
In essence, the task of salient object detection is related to
these three levels at the same time.
Therefore, how to effectively model
all the above factors in a unified learning framework is a key and challenging issue
to solve in salient object detection.
\begin{figure}[t]
    \centering 
        \includegraphics[width=1\linewidth]{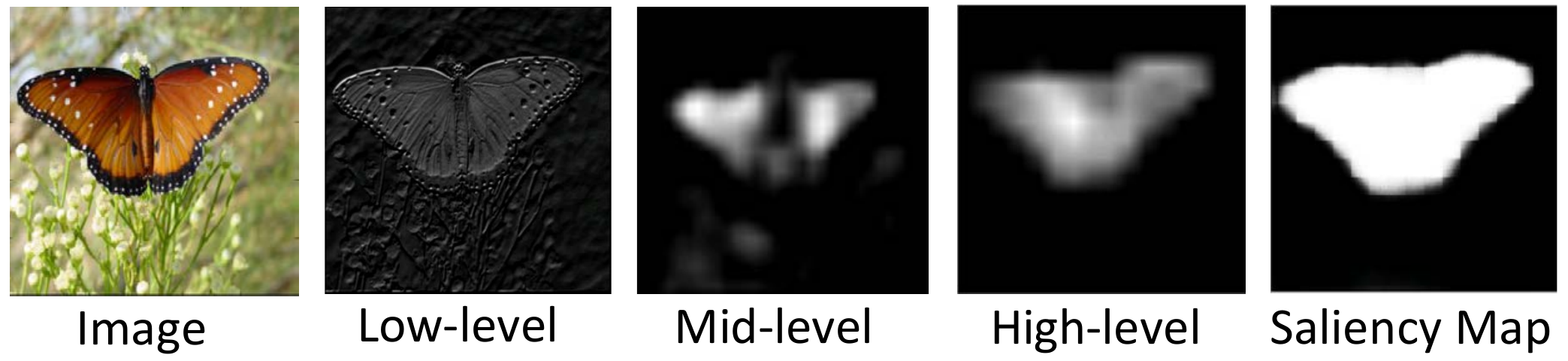}
    \centering
    \caption{Illustration of multi-level semantic information
        obtained by FCNN. FCNN captures the saliency properties
        across different levels (i.e., low-level (image edge information), mid-level (coarse-grained shape), and high-level (semantic object information)).
    }
    \label{fig:level}
    \vspace{-1.5\baselineskip}
\end{figure}
Most existing methods use various prior knowledge to detect salient objects, such as background priors~\cite{GMR, RBD,BSCA2015}, center priors~\cite{HS} and contrast priors~\cite{CenterSurround}. However, these models have to
be carefully tailored for adapting to different types of image data with
a wide variety of objects and their contextual interactions, thereby
making them less applicable to a wide range of problems in practice.
For example, salient objects often appear at
the image boundary (far away from the image center), and thus the
center prior is less useful.
Moreover, as parts of salient objects may be similar to some background
regions, it is difficult to detect them only based on the contrast information.

Motivated by these observations, we construct an adaptive model to effectively capture the intrinsic semantic
properties of salient objects and their essential differences
from the background in a pure data-driven framework.
Meanwhile, the model is supposed to encode the
saliency information across different levels (from low-level
to high-level), as shown in Figure~\ref{fig:level}. To address these problems, a number of deep learning
approaches~\cite{AlexNet, VGGnet, GoogLeNet,FCN} have emerged as
a powerful tool of data-driven
multi-granularity image understanding. Therefore, how to
design an effective deep learning model for saliency detection
is the focus of this work.

In this paper, we propose a multi-task deep saliency model based on a fully convolutional
neural network (FCNN) with global input (whole raw images) and global
output (whole saliency maps).
In principle, the deep saliency model
takes a data-driven learning pipeline
for capturing the underlying saliency prior knowledge, and subsequently
builds
a multi-task learning scheme for
exploring
the intrinsic correlations
between the tasks of saliency detection
and semantic image segmentation, which share
the fully convolutional layers in the learning process of FCNN.
Through collaborative feature learning from such two correlated tasks,
discriminative features are extracted
to effectively encode the object perception information.
Moreover, the deep saliency model has the capability of
capturing
the semantic information on salient objects across different levels
using FCNN, which
explores
the feature-sharing properties of salient objects.
In addition, we further develop a graph
Laplacian regularized nonlinear regression scheme for saliency
refinement to generate a fine-grained boundary-preserving saliency map.

The main contributions of this work are summarized as follows:

1) We propose a multi-task FCNN based approach to model the
intrinsic semantic properties of salient objects in a totally data-driven
manner. The proposed approach performs collaborative feature learning
for the two correlated tasks (i.e., saliency detection
and semantic image segmentation), which generally leads to
the performance improvement of saliency detection in object perception.
Moreover, it effectively enhances the feature-sharing capability of
salient object detection by using the fully convolutional layers, resulting in
a significant reduction of feature redundancy.

2) We present a fine-grained super-pixel driven saliency refinement
model based on graph Laplacian regularized nonlinear regression with
the output of the proposed FCNN model.
The presented model admits a closed-form solution,
and is able to accurately preserve object boundary information for
saliency detection.

\section{Related Work}
\subsection{Saliency Detection Approaches}~\label{sec:related_saliency}
Saliency models have been built for visual attention modeling~\cite{IT98,GB,huang2015salicon,DeepGaze14} and salient object detection~\cite{FT,GMR,Zhao2015}. The former task aims to predict human fixation locations on natural images, while the proposed method aims to compute the pixel-wise saliency values for capturing the regions of the salient objects. A more detailed survey about saliency models can be found in~\cite{Survey}.

Early approaches in this field typically formulate
salient object detection as the problem of image contrast analysis in
a center-surround or object-background manner.
These approaches detect the salient objects by computing the local
center-surround differences~\cite{IT98,Contrast03, CenterSurround}, or evaluating
the region uniqueness and rarity in the scene~\cite{MSRA, Hypergraph, ChengPAMI15}.
Furthermore, a number of approaches heuristically enforce some predefined priors
on the process of saliency detection such as image center priors, large-sized
closed regions~\cite{HS}, and semantic object priors~\cite{LowRank}.
In addition, several approaches~\cite{GS, GMR, RBD,BSCA2015} attempt to utilize
the background prior information to perform the label propagation
procedure in a spatially structural graph. Instead of manually assuming these priors in advance, we automatically learn the prior knowledge in a totally data-driven manner.

\begin{figure}[t]
	\centering 
	\includegraphics[width=1\linewidth]{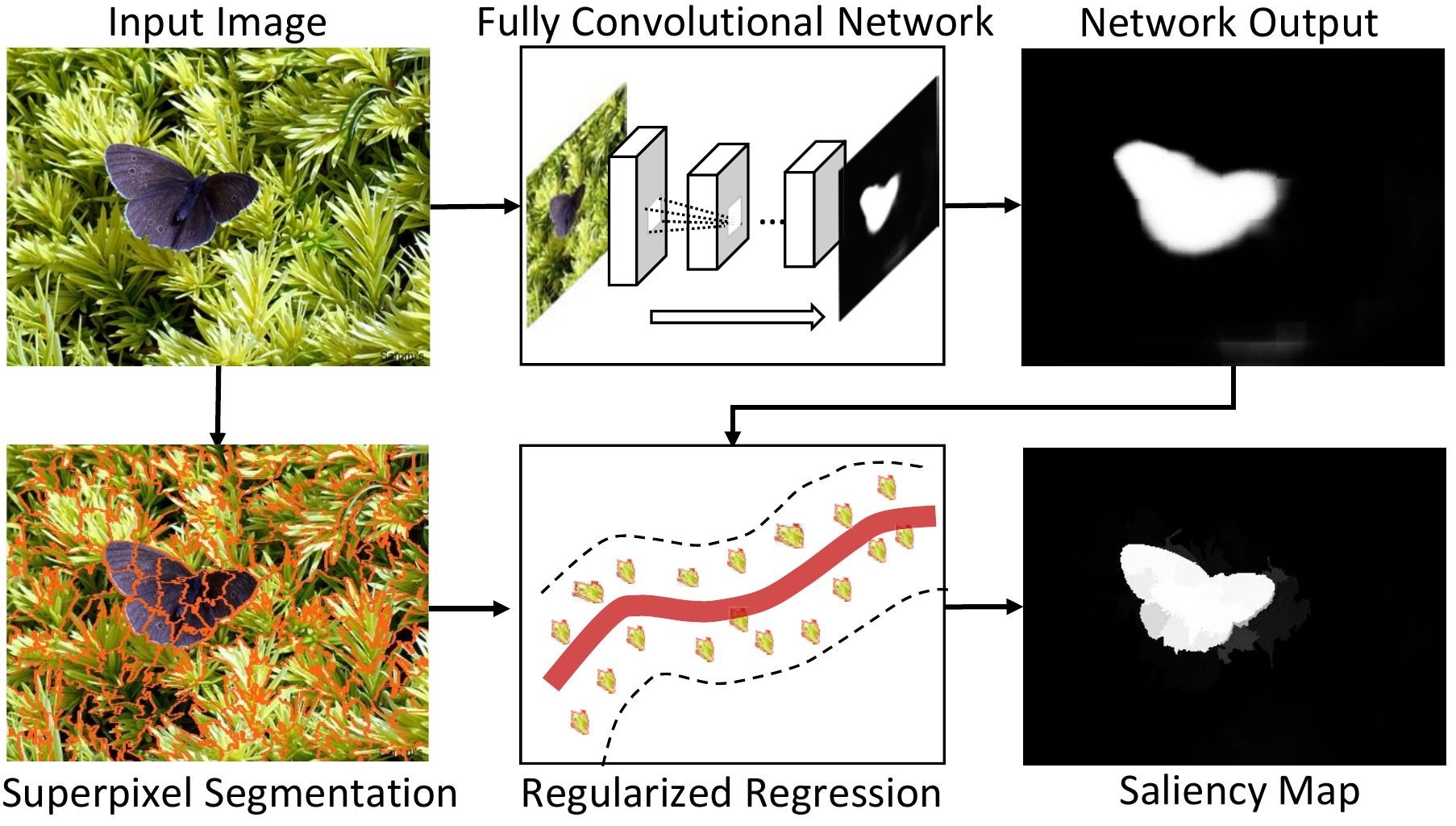}
	\centering 
	\caption{Illustration of our approach for salient object
		detection.
		First, a fully convolutional neural network takes the whole image as
		input and predicts the saliency map by capturing the semantic
		information on salient objects across different levels.
		Second, a Laplacian regularized nonlinear regression scheme based on
		the super-pixel graph is used to produce a fine-grained
		boundary-preserving saliency map.
	}
	\label{fig:flow}
	\vspace{-1.0\baselineskip}
\end{figure}

\subsection{Deep Saliency Networks}~\label{sec:related_deep}
At present, deep neural networks have been applied to detect salient objects~\cite{LEGS, SalDeepFeature,Zhao2015,li2015visual}.
These data-driven saliency models, including~\cite{DRFI}, aim to directly
capture the semantic properties of salient objects in terms of
supervised learning from a collection of training data with pixel-wise
saliency annotations.
By independently modeling all local image patches
(treated as training samples for classification or regression), these approaches are often incapable of effectively capturing the feature-sharing
properties of salient objects, leading to a high
feature redundancy as well as an expensive computational cost. Furthermore,
they mainly focus on learning a saliency prediction function to model the relative information between the objects regardless of what they are, thus lacking the capability of
effectively modeling the object perception properties of
saliency detection (i.e., objectness).
In practice, such object perception properties are very helpful
to discovering salient objects.

%
%
\begin{figure*}[t]
    \centering 
    \includegraphics[width=1\textwidth]{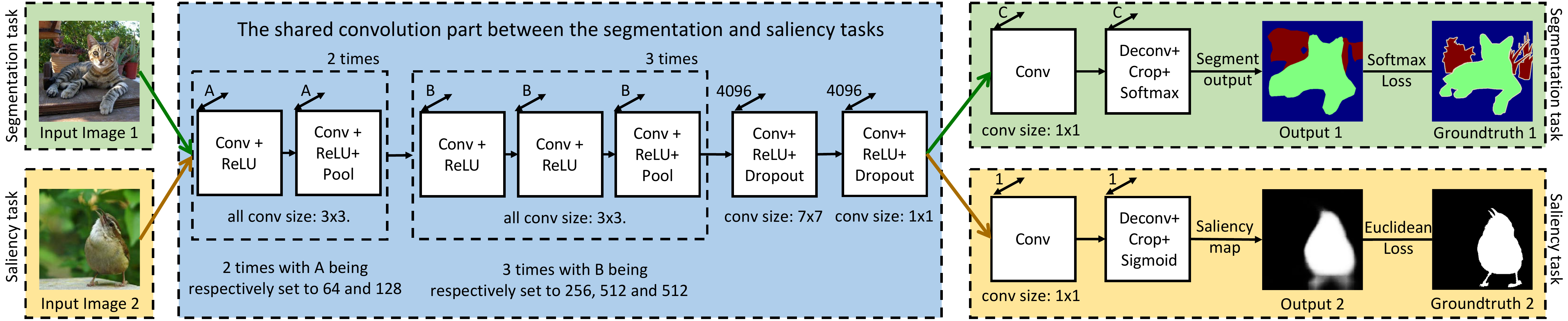}
    \centering
    \caption{Architecture of the proposed fully convolutional neural
        network for training.
        The FCNN scheme carries out the task of saliency detection
        in conjunction with the task of object class segmentation, which share
        a convolution part with $15$ layers.
        The segmentation task seeks for the intrinsic
        object semantic information of the image,
        while the saliency task aims to find the salient objects. For the saliency detection task, we only use the saliency-related network for testing.
    }
    \label{fig:net}
    \vspace{-1.5\baselineskip}
\end{figure*}

\subsection{Multi-task Neural Networks}~\label{sec:related_multitask}
Designing deep architectures for joint tasks is popular and effective, which has been used for several vision tasks~\cite{Eigen2015,Pinheiro2015,girshick2015fast,ren2015faster}. The networks in~\cite{Eigen2015} are trained separately for depth prediction, surface normal estimation, and semantic labeling tasks. The configurations of these three networks are similar, but the learned parameters are totally different.
All of the methods in~\cite{Pinheiro2015,girshick2015fast,ren2015faster} solve a common problem with two objective terms, such as generating object segmentation proposals with segmenting and scoring~\cite{Pinheiro2015}. In contrast, we use two networks performing the segmentation and saliency detection tasks by sharing features, which forms a tree-structured network architecture. To our knowledge, it is novel to introduce the semantic segmentation task into the deep nets to learn a better saliency model.

\section{Proposed Approach}

\subsection{Overview}

The proposed salient object detection approach mainly consists of two
components:
1) multi-task fully convolutional neural network (FCNN); and 2) nonlinear regression
for saliency refinement.
For 1), we perform the task of learning a FCNN saliency model in conjunction
with the task of pixel-wise object class segmentation.
These two tasks share a set of fully convolutional layers for extracting
the features of a given image on multiple levels, and meanwhile have different deconvolutional layers that are tailored for different applications, including
saliency detection and object class segmentation.
Such a multi-task learning scheme enables saliency detection to have a
better object perception capability.
For 2), based on the output saliency results from FCNN, we further
refine the saliency map by graph Laplacian regularized nonlinear regression,
thereby generating fine-grained boundary-preserving saliency results.

Given an image, we first use the FCNN to compute a coarse-grained saliency map as foreground information. Meanwhile, boundary super-pixels of the image are treated as background seed samples~\cite{GS, GMR, RBD,BSCA2015}, and then another coarse-grained saliency map can be computed by the non-linear regression based propagation. After that, the coarse-grained foreground and background saliency maps are combined and finally refined based on graph Laplacian regularized nonlinear regression.
Figure~\ref{fig:flow} shows the main steps of the proposed salient object detection approach.
We will detail the descriptions of multi-task FCNN, Laplacian regularized regression and the inference process in the following subsections.

\subsection{Multi-Task FCNN}

\subsubsection{Network Architecture}
As illustrated in Figure~\ref{fig:net},
our FCNN scheme carries out the task of saliency detection
in conjunction with the task of object class segmentation.
%
More specifically, the convolutional layers of our FCNN between the
object segmentation and saliency detection tasks are shared.

The shared convolution part aims to extract a collection of features for the input image across
different semantic levels.
This part processes the input RGB image by a sequence of convolutional operations
across $15$ convolutional layers, each of which is equipped with a
Rectified Linear Unit(ReLU)~\cite{ReLU}.
In addition,
some layers are followed by the max pooling operations, as shown in
Figure~\ref{fig:net}.
In practice, the first $13$ layers are initialized from the VGG
nets~\cite{VGGnet} (pre-trained over the ImageNet dataset~\cite{ImageNet}
with a large number of semantic object classes), resulting in the
discriminative power of our shared convolution part for semantic
object classification.
To further model the spatial correlations of the whole
image,
we only use the fully convolutional learning architecture
(without any fully connected layers).
That is because the fully convolutional operations have the capability of
sharing the convolutional
features across the entire image, leading to the
reduction of feature redundancy. Therefore, the fully convolutional learning architecture
is simple and efficient with global input and global output.

\subsubsection{Object Segmentation Task}
In the segmentation task, features extracted from
the shared  convolution part are fed into the semantic segmentation
task, which seeks for the intrinsic object semantic information for the
image.
We first apply one layer with $1\times1$ sized convolution to
compute the segmentation score map.
To make the output map have the same size as the input image,  we need a
deconvolution layer, which is also used in the FCN net~\cite{FCN}.
The parameters in the deconvolution layer are updated during training
to learn an up-sampling function.
The up-sampled outputs are then cropped to the same size of the input
image.
Thus, the network takes the whole image as input and produces the
corresponding pixel-wise dense prediction results with the same size,
and thereby preserves the global information of the image.
The segmentation task finally outputs $C$ probability maps for $C$
object classes (including the background),
and can be trained with a pixel-wise softmax loss against the
ground-truth segmentation map.

\subsubsection{Saliency Detection Task}
In the saliency task, our network aims to find the interesting objects
from an image.
Specifically, we use one convolution layer and one
deconvolution layer to generate the saliency map (normalized to $[0,1]$
by the sigmoid function).
Finally, the saliency network ends up with a
squared Euclidean loss layer for saliency regression, which
learns the saliency properties from data.

\subsubsection{Multi-Task Learning}
We perform FCNN learning by
taking into account the saliency detection
task in conjunction with the
segmentation task.
Let $\mathcal{X}= \{\X_{i}\}_{i=1}^{N_1}$ denote a collection
of training images (with
the width and height of each image
being respectively $P$ and $Q$),
$\{\Y_{ijk}|\Y_{ijk} \in \{1,2,\dots,C\} \}_{N_1\times{P}\times{Q}}$
denote
their corresponding pixel-wise ground truth  segmentation maps;
and $\mathcal{Z}= \{\Z_{i}\}_{i=1}^{N_2}$ denote a set of training images with
the corresponding ground-truth binary map of salient objects
being $\{\M_i\}_{i=1}^{N_2}$.
Furthermore, we denote all the parameters in the shared convolution
part as $\seita_s$;
the parameters in the segmentation task as $\seita_h$;
and the parameters in the saliency task as $\seita_f$.
Our FCNN is trained by minimizing the following
cost functions:
\begin{equation}\label{eq:net}
\begin{aligned}
&\J_{1}(\mathcal{X};\seita_s,\seita_h)=\\
&-\frac{1}{N_1}\sum_{i=1}^{N_1}
\sum_{c=1}^C\sum_{j=1}^{P}\sum_{k=1}^{Q}{\one\{\Y_{ijk}=c\}\log(h_{cjk}(\X_i;\seita_s,\seita_h))};\\
&\J_2(\mathcal{Z};\seita_s,\seita_f)=\frac{1}{N_2}\sum_{i=1}^{N_2}
\|\M_i-f(\Z_i;\seita_s,\seita_f)\|_F^2
\end{aligned}
\end{equation}
where $\one$ is the indicator function;
$h$ is a semantic segmentation function returning $C$ probabilistic segmentation maps and $h_{cjk}$ is the $(j,k)$-th element of the $c$-th
probabilistic segmentation map;
and $f$ is the saliency map output function.
Clearly, the first one is associated with
the cross-entropy loss term for the segmentation task,
while the second cost function corresponds to
the squared Euclidean loss term for the saliency task.

Subsequently, we train the network by the stochastic gradient descent
(SGD) method to minimize the above cost functions with regularization
on all the training samples.
In practice,
we carry out the segmentation task and the saliency task in an alternating manner.
Namely,  we first learn the parameters $\seita_s$ and $\seita_h$
for the segmentation task, and then learn the parameters $\seita_f$
as well as update the parameters $\seita_s$
for the saliency task.
The above procedure is iteratively repeated as training proceeds.

\begin{figure}[t]
\centering 
\includegraphics[width=1\linewidth]{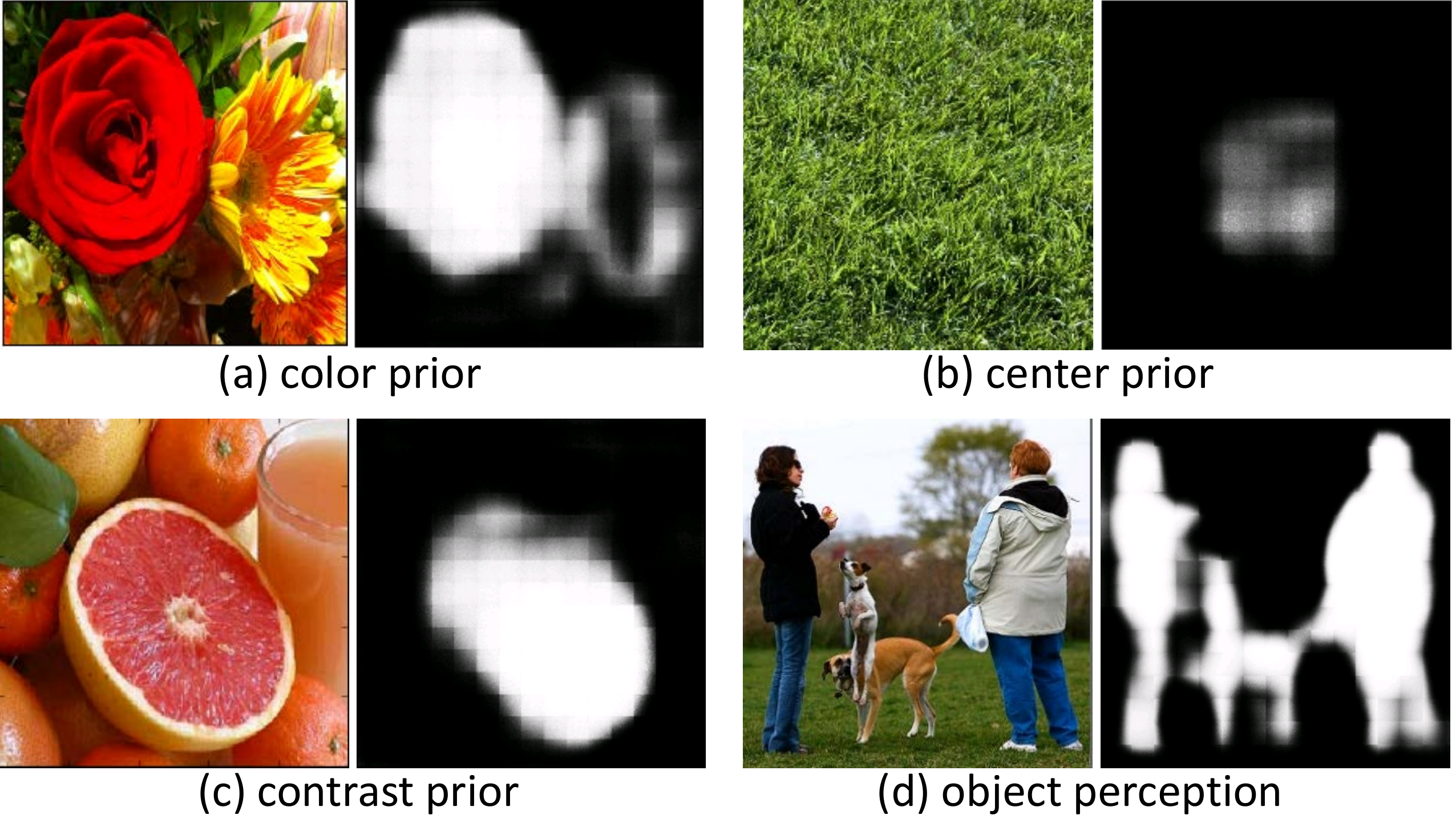}
\centering
\caption{Some intuitive examples of the saliency properties learned
 in our network, which are consistent with human perception.
Given an image, people tend to pay their attention to red objects \cite{LowRank,Itti_Nature} (a),
the image center \cite{MSRA} (b), regions with high contrast in texture and shape\cite{CenterSurround} (c), and semantic objects \cite{SVO} (d).
}
\label{fig:prior}
\vspace{-1.5\baselineskip}
\end{figure}

We present some intuitive examples to show that our network captures
the underlying  saliency properties in Figure~\ref{fig:prior}.
Specifically, the saliency value of the red flower in
Figure~\ref{fig:prior}(a) is higher than  the yellow one,
which is consistent with human perception (paying more attention to
the objects with red bright colors \cite{LowRank,Itti_Nature}).
Given an image with only the background (e.g., the lawn shown in
Figure~\ref{fig:prior}(b)), people often put their eyes at the center
of the image\cite{MSRA}.
As shown in Figure~\ref{fig:prior}(c), our network is able to
detect objects with a high contrast to the surroundings \cite{CenterSurround} (i.e., the
salient orange slice is distinct from the surrounding oranges in shape and texture).
With the help from the semantic segmentation learning task, our
network is able to detect the semantic objects in the scene, as shown in Figure~\ref{fig:prior}(d).
%
%
%
\subsection{Regularized Regression for Refinement}
As shown in Figure~\ref{fig:prior},
the saliency maps generated from the fully convolutional neural network
are usually with fuzzy object boundaries.
In order to
well preserve the object boundaries for the saliency maps, we make use of
internally homogeneous and boundary-preserving super-pixels (obtained
by over-segmentation using SLIC~\cite{SLIC}) as basic representation
units, and then construct a super-pixel level adjacency graph
to model the topological relationships among super-pixels in both
spatial and feature dimensions.
More specifically, a given image is first over-segmented into a set of super-pixels
$\{{\bf{x}}_{i}\}_{i=1}^{N}$, each of which is represented by a Lab
color  vector (obtained by averaging the pixel-wise color features
within the super-pixel).
Therefore, the super-pixel level graph with the affinity matrix
$W=(w_{ij})_{N\times N}$ is constructed as follows:
\begin{equation}\label{eq:W}
w_{ij}=
\left\{
\begin{array}{ll}
K({\bf{x}}_{i},{\bf{x}}_{j}) & \mbox{if} \thinspace {\bf{x}}_{i} \thinspace \mbox{and} \thinspace {{\bf{x}}_{j}} \thinspace \mbox{are spatially adjacent}
\\
0 & \mbox{Otherwise}
\end{array}
\right.
\end{equation}
where $K({\bf{x}}_{i},{\bf{x}}_{j})$ stands for the RBF kernel evaluating the
feature similarity (such that
$K({\bf{x}}_{i},{\bf{x}}_{j}) = \exp(-\frac{1}{\rho}\|{\bf{x}}_{i} -
{\bf{x}}_{j}\|_{2}^{2})$).
Let ${\bf{y}} = (y_{1}, \ldots, y_{N})^{\top}$
denote the pre-specified saliency score vector (ranging from [-1,1])
corresponding to the super-pixels $\{{\bf{x}}_{i}\}_{i=1}^{N}$.
In terms of graph-based semi-supervised regression, some super-pixels are treated as seed samples with pre-specified saliency scores (such as boundary super-pixels), and the saliency states for the remaining super-pixels are temporarily undetermined (initialized to be 0) until they are reached by propagation.
Without loss of generality, we assume the first $l$ super-pixels
have the initial saliency information while the last $u$ super-pixels
are null (such that $N = l + u$).
Hence, our task is to learn a nonlinear regression
function for saliency prediction over a given super-pixel ${\bf{x}}$ (such that $g({\bf{x}})
= \sum_{j=1}^{N}\alpha_{j}K({\bf{x}}, {\bf{x}}_{j})$) within the following
optimization framework:
\begin{equation}\label{eq:loss}
\begin{aligned}
\min_{g}\frac{1}{l}\sum_{i=1}^{l}(y_{i}-g({\bf{x_{i}}}))^{2} + \gamma_{A}\|g\|_{K}^{2} + \frac{\gamma_{I}}{(l+u)^{2}}\mathbf{g}^{\top}\mathbf{L}\mathbf{g}
\end{aligned}
\end{equation}
where $\|g\|_{K}$ denotes the norm of $g$ in the RKHS~\cite{RKHS} induced by the
kernel function $K$;
$\mathbf{g} = (g({\bf{x_{1}}}),\ldots, g({\bf{x_{N}}}))^{\top}$;
$\mathbf{L}$ stands for the graph Laplacian
matrix for the affinity matrix $W$;
and $\gamma_{A}$ as well as $\gamma_{I}$
are two trade-off control factors ($\gamma_{A}=10^{-6}$ and
$\gamma_{I}=1$ in the experiments).
Clearly, the first term in the above optimization problem corresponds to the squared regression loss, and
the third term ensures the spatial smoothness of the final saliency
map.
The above optimization problem can be equivalently transformed to:
\begin{equation}\label{eq:alpha}
\begin{aligned}
\min_{\boldsymbol{\alpha}}\frac{1}{l}\|{\mathbf{y}}-\mathbf{J}\mathbf{K}\boldsymbol{\alpha}\|_{2}^{2}
+\gamma_{A}\boldsymbol{\alpha}^{\top}\mathbf{K}\boldsymbol{\alpha}
+\frac{\gamma_{I}}{(l+u)^{2}}\boldsymbol{\alpha}^{\top}\mathbf{K}
\mathbf{L}\mathbf{K}\boldsymbol{\alpha}
\end{aligned}
\end{equation}
where $\mathbf{J} = \mbox{diag}(\overbrace{1,\ldots,1}^{l},
\overbrace{0,\ldots,0}^{u})$  is a diagonal matrix,
$\boldsymbol{\alpha}$ corresponds to the nonlinear regression
coefficient vector $(\alpha_{i})_{i=1}^{N}$,
and $\mathbf{K}=(K({\bf{x}}_{i},{\bf{x}}_{j}))_{N\times N}$ is the kernel Gram matrix.
The optimal solution to the above optimization
problem is formulated as:
\begin{equation}\label{eq:finalAlpha}
\begin{aligned}
\boldsymbol{\alpha}^{\ast}=\left(\mathbf{J}\mathbf{K}+\gamma_{A}l\mathbf{I}
+\frac{\gamma_{I}l}{(l+u)^{2}}\mathbf{L}\mathbf{K}\right)^{-1}\mathbf{y}
\end{aligned}
\end{equation}
where $\mathbf{I}$ is the identity matrix. As a result, we have the nonlinear regression function
$g({\bf{x}})
= \sum_{i=1}^{N}\alpha_{i}^{\ast}K({\bf{x}}, {\bf{x}}_{i})$
with $\alpha_{i}^{\ast}$ being the $i$-th element of
$\boldsymbol{\alpha}^{\ast}$. Based on $g({\bf{x}})$,
we can compute the saliency score for any given super-pixel
$\bf{x}$ within an image.

\subsection{Generating Saliency Map}
\begin{figure}[t]
\centering 
\includegraphics[width=1\linewidth]{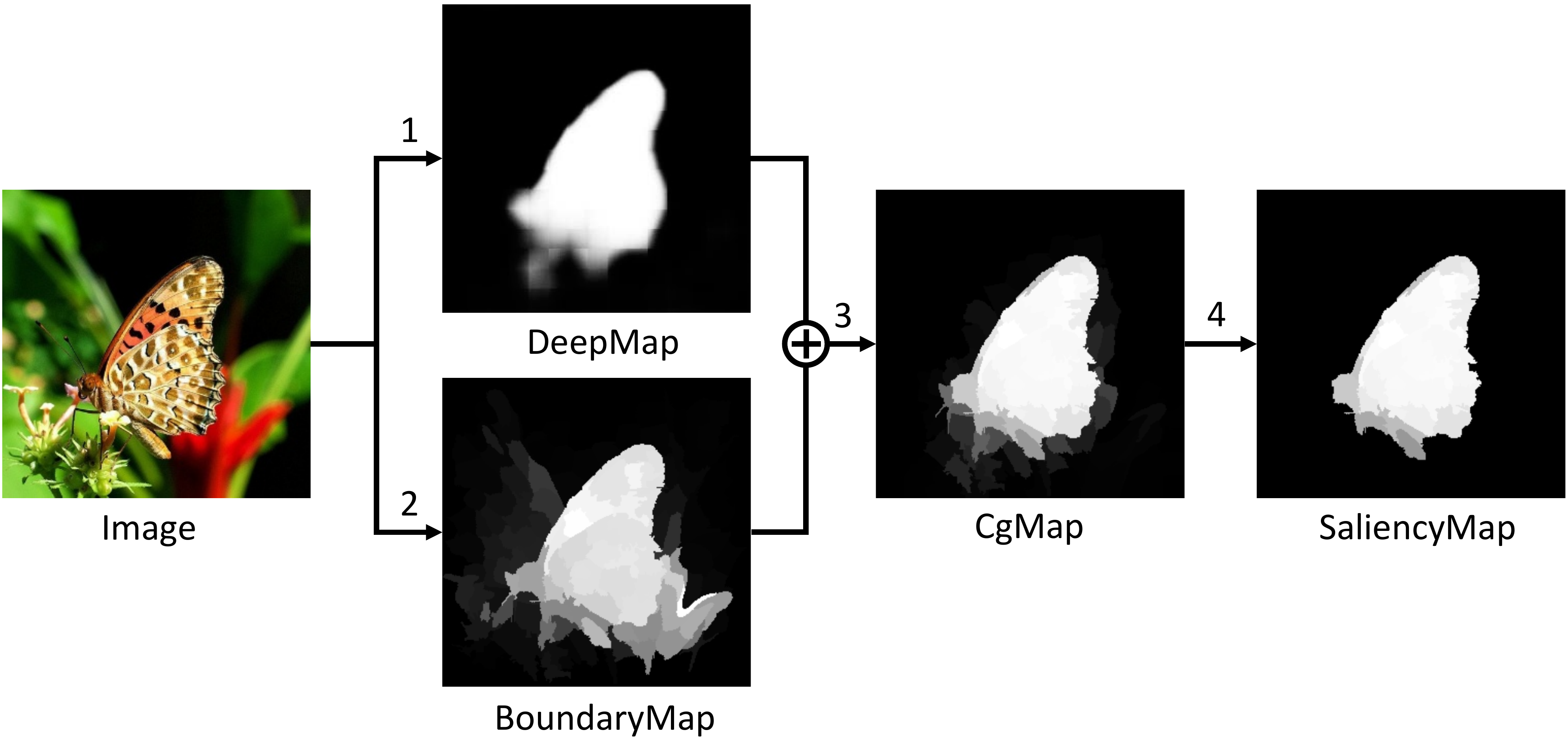}
\centering
   \caption{Generating a saliency map with four stages.
    Clearly, our saliency detection
    approach obtains a visually precise saliency map.
    }
\label{fig:computation}
\vspace{-1.5\baselineskip}
\end{figure}

Given an image, the saliency map is computed in four stages:
1) object perception by FCNN;
2) image boundary information propagation within the super-pixel graph;
3) coarse-grained saliency information fusion; and
4) fine-grained saliency map generation by
nonlinear regression-based propagation, as illustrated in
Figure~\ref{fig:computation}.

For 1), the trained FCNN is used to adaptively capture the
semantic structural information on object perception,
resulting in a pixel-wise objectness probability map (ranging from 0
and 1), which we refer as a {\it DeepMap}.
This stage focuses on modeling the underlying
object properties from the perspective of foreground discovery using FCNN.

In contrast, the stage 2) aims to explore the influence of
the image boundary information in saliency detection
from the viewpoint of background propagation.
%
%
Namely, we use the learned regression function (defined in Eq.~\ref{eq:alpha}) based
on a Laplacian graph
to estimate the saliency values on the super-pixel level
where the ones on the image boundary are initialized -1 and others as 0.
After the propagation process, we have a saliency map
denoted as {\it BoundaryMap} (normalized to $[0,1]$).

In stage 3), we perform saliency fusion of the {\it DeepMap} and {\it BoundaryMap}
to generate the coarse-grained saliency map
(referred to as  {\it CgMap}) by
\begin{equation}
CgMap = DeepMap^{1-\beta}\circ BoundaryMap^{\beta}
\end{equation}
where $\beta$ is a trade-off control factor and
$\circ$ is the elementwise product operator.

In stage 4), the normalized $CgMap$ (s.t. $[-1,1]$) is
fed into Eq.~\ref{eq:alpha} for saliency refinement
over the super-pixel graph,
resulting in the final fine-grained saliency map.
As shown in
Figure~\ref{fig:computation},
based on graph Laplacian regularized nonlinear regression,
our saliency detection approach is able to obtain more accurate saliency detection
results with fine-grained object boundaries.
%
%
%
\section{Experimental Results}

\subsection{Experimental Setup}\label{sec:setup}

\subsubsection{Datasets}
In order to evaluate the performance of the proposed approach, we
conduct a set of qualitative and quantitative experiments on eight
benchmark datasets annotated with pixel-wise ground-truth labeling,
including the
ASD~\cite{FT}, DUT-OMRON~\cite{GMR}, ECSSD~\cite{HS},
PASCAL-S~\cite{PASCAL-S},  THUR~\cite{THUR}, THUS~\cite{ChengPAMI15},
SED2~\cite{SED2}, and SOD~\cite{SOD} datasets.
Specifically, the first six datasets are composed of object-centric
images distributed in relatively simple scenes (e.g., high object
contrast and clean background).
In contrast, the last two datasets are more challenging with multiple
salient objects and background clutters in images.
For descriptive convenience, Table 1 shows the
basic information of these eight datasets.
\begin{table}[htbp]
  \centering
    \resizebox{1\linewidth}{!}{
    \begin{tabular}{|c|c|c|c|c|c|c|c|c|}
    \hline
    Dataset& ASD   & THUS  & THUR  & DUT-OMRON & ECSSD & PASCAL-S & SOD   & SED2 \\
   \hline
    Size & 1000  & 10000 & 6232  & 5168  & 1000  & 850   & 300   & 100 \\
    \hline
    \end{tabular}
    }
  \caption{Illustration of the image numbers associated with the eight datasets.}
  \label{tab:datasets}

  \vspace{-1.0\baselineskip}
\end{table}%
More specifically, ASD is a commonly used dataset in salient object detection.
For each image in ASD, there is usually one dominant salient object with simple
backgrounds. Compared with ASD, THUS has a larger dataset size,
and covers more saliency cases. It is noted that ASD is a subset of THUS.
Moreover, THUR is generated by crawling from Flick, and consists of
the images with 5 object classes, including
``butterfly'', ``coffee mug'', ``dog jump'', ``giraffe'', and ``plane''.
Among these images, there are 6232 images that have pixel-wise saliency annotation
maps used for saliency detection. Aiming at  overcoming
the drawbacks of ASD (i.e., limited
objects and simple background),
DUT-OMRON is manually selected from
 more than
140,000 natural images, each of which has one or more
salient objects and relatively complex backgrounds.
As an extension of the Complex Scene Saliency Dataset(CSSD)~\cite{HS},
ECSSD is obtained by aggregating the images from the two
publicly available datasets (i.e., BSD~\cite{BSD}
and PASCAL VOC~\cite{VOC}) and the internet.
PASCAL-S is generated from the PASCAL VOC dataset~\cite{VOC}
with 20 object categories and complex scenes.
SED2 is a multi-saliency-object dataset that usually includes two salient objects
in each image.
SOD is composed of the images (based on the BSD dataset~\cite{BSD})
with one or more objects and complex backgrounds.
A few images in SOD are also part of ECSSD.

\subsubsection{Implementation details}
In the experiments, our algorithm is implemented in MATLAB on a desktop
computer with an Intel E5-2609 CPU(2.4 GHz) and 8 GB RAM.
During the saliency regression process, the super-pixel oversegmentation is
carried out by the SLIC method~\cite{SLIC} with the super-pixel number
$N$ being 200.
The scaling factor $\rho$ in the RBF kernel is set to $0.1$.

The fully convolutional neural network (FCNN) is implemented
on the basis of the Caffe~\cite{Caffe} toolbox.
More specifically,
we initialize the first $13$ convolutional layers of FCNN
with those of the pretrained
VGG $16$-layer net~\cite{VGGnet} and transfer the learned
representations by fine-tuning~\cite{DeCAF} to the semantic segmentation task
and the saliency detection task.
We construct the deconvolution layers
by upsampling, whose parameters are initialized
as simple bilinear interpolation parameters
and iteratively updated during training. We resize all the images and ground-truth maps to $500\times500$ pixels for training,
the momentum parameter is chosen as $0.99$, the
learning rate is set to $10^{-10}$, and the weight decay
is $0.0005$. The SGD learning procedure is
accelerated using a NVIDIA Tesla K40C GPU device,
and takes
around $3$ days in $80,000$ iterations.

In the experiments, the segmentation task
and the saliency detection task
in the FCNN are optimized in an alternating manner, since none of the existing datasets contain both segmentation and saliency annotations. We will show that our method can transfer the segmentation knowledge into the saliency detection for learning a better feature representation, without recollecting the needed training data. Note that our method is flexible and can be trained jointly if such datasets are available. During the training process, both two task-related networks share the parameters $\boldsymbol{\theta}_{s}$ of the shared convolution part.
Namely, we firstly perform the segmentation task using
the object class segmentation dataset (i.e., PASCAL VOC 2007 with the
number of object  classes $C=21$)
to obtain $\boldsymbol{\theta}_{s}$
and the segmentation-related network parameters
$\boldsymbol{\theta}_{h}$. More specifically, the detailed training procedure is carried out as follows:
\begin{enumerate}

\item Initialize the parameters $\seita_s^{0}$ of the shared fully convolutional part using the pretrained VGG 16-layer net.

\item Initialize the parameters $\seita_h^{0}$ (for the segmentation task)
     and $\seita_f^{0}$ (for the saliency task) randomly from the normal distribution.

\item Based on $\seita_s^{0}$ and $\seita_h^{0}$, utilize SGD to train the segmentation-related net for updating these two parameters, resulting in
     $\seita_s^{1}$ and $\seita_h^{1}$.

\item Using $\seita_s^{1}$ and  $\seita_f^{0}$,  perform SGD to train the saliency-related net for updating the saliency-related parameters,
     resulting in $\seita_s^{2}$ and  $\seita_f^{1}$.

\item Based on $\seita_s^{2}$ and $\seita_h^{1}$, use SGD to train the segmentation-related net for obtaining
     $\seita_s^{3}$ and $\seita_h^{2}$.

\item Using $\seita_s^{3}$ and  $\seita_f^{1}$,  perform SGD to train the saliency-related net for updating the saliency-related parameters,
     resulting in $\seita_s^{4}$ and  $\seita_f^{2}$.

\item Repeat the above steps (3)-(6) until obtaining the final parameters $\seita_s$, $\seita_h$ and $\seita_f$. In practice, the alternating training procedure is repeated for three times in the experiments, often leading to a relatively stable performance.
\end{enumerate}

We note that the above experimental configurations are fixed throughout
all the experiments.

\subsection{Evaluation Metrics}
In the experiments, we utilize four metrics for quantitative performance
evaluations following~\cite{Benchmark}, including Precision and Recall (PR) curve, F-measure,
mean absolute error (MAE), ROC curve and area under ROC curve (AUC).
Specifically, the PR curve reflects the
object retrieval performance in precision and recall by binarizing the
final saliency map using different thresholds (usually ranging from 0
to 255).
The F-measure characterizes the balance degree of object retrieval
between precision and recall such that
$F_{\eta}=\frac{(1+\eta^{2}) Precision \times Recall}{\eta^{2}\times Precision + Recall}$,
where $\eta^{2}$ is typically set to 0.3 like the most existing literature work,
The Precision and Recall rates correspond to the object retrieval
performance after binarization using a particular threshold.

Typically, maximum F-measure (maxF) is associated with the maximum
F-measure value
computed from the PR curve (as suggested in~\cite{maxF}), while
average F-measure  (aveF) uses the adaptive threshold
(i.e., twice the average value of the final saliency map~\cite{FT}) for
binarization.
In addition, MAE refers to the average pixel-wise error between the saliency map
and ground truth.
The ROC curves are plotted w.r.t. false positive rate (FPR) and true positive rate (TPR), which are defined as $FPR=\frac{M\cap \bar{G}}{\bar{G}}$, $TPR=\frac{M\cap G}{G}$,
where $M$ denotes the binary mask of the saliency map with a particular binarization threshold,
$G$ denotes the ground-truth binary map, and $\bar{G}$ denotes the opposite of $G$.
In the experiments, the ROC curves are generated by varying the binarization thresholds
from 0 to 255 over the final saliency maps.

Finally, AUC evaluates the object detection performance,
and computes the area under the standard ROC curve (w.r.t. false positive rate
and true positive rate).

\subsection{State-of-the-art performance comparison}
We qualitatively and quantitatively
compare the proposed approach with several state-of-the-art
methods including DeepMC~\cite{Zhao2015}, DRFI~\cite{DRFI}, Wco~\cite{RBD}, GC~\cite{GC},
GMR~\cite{GMR},  FT~\cite{FT}, MC~\cite{MC},
HS~\cite{HS}, SVO~\cite{SVO}, DSR~\cite{DSR}, LEGS~\cite{LEGS}, BL15~\cite{BL} and BSCA~\cite{BSCA2015}.
Among these methods, DeepMC (deep learning), LEGS (deep learning) and DRFI (random forest regressor)£¬ are also learning-based methods; MC and DSR (background prior), SVO (objectness + visual saliency), HS (global contrast), Wco, GMR and MC (boundary prior), BL15 (center and color priors + bootstrap learning) are state-of-the-art salient object detection models which are designed with different assumptions; the classic methods FT are included as a baseline.
Most of the saliency maps associated with the
competing approaches can be obtained by running their publicly
available source code using the default experimental
configurations.
For the LEGS, BL15 and BSCA methods, we use the results reported in the
literature.
%
%
%
\begin{figure*}[htbp]
\centering 
\includegraphics[width=1\textwidth]{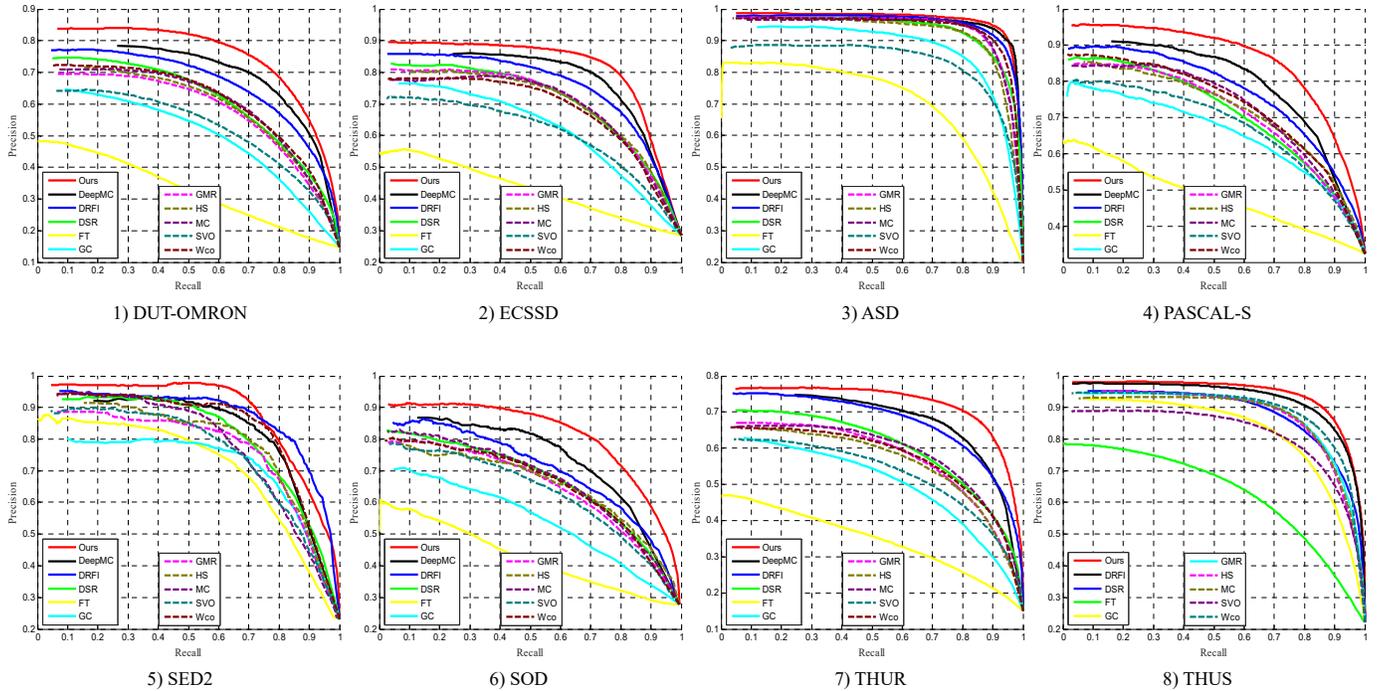}
\centering
\caption{Precision-recall curves of different saliency detection
    methods on $8$ benchmark datasets.
    Overall, the proposed approach performs well with higher precision
    in the case of a fixed recall.
}
\label{fig:prCurve}
\vspace{-1.5\baselineskip}
\end{figure*}
%
%
%
\begin{table*}[htbp] \scriptsize 
\renewcommand{\tabcolsep}{2pt}
\centering
\resizebox{1\textwidth}{!}{
    \begin{tabular}{|c|c|c|c|c|c|c|c|c|c|c|c|c|c|c|c|c|}
        \hline
        Dataset &       & OurLO & OurBR & DeepMC & DRFI  & Wco   & DSR   & GC    & GMR   & HS    & MC    & SVO   & FT    & LEGS  & BL15  & BSCA \\
        \hline
        \multirow{4}[1]{*}{DUT-OMRON} & aveF  & \second{0.6045} & 0.5947 & \best{0.6209} & 0.5519 & 0.5276 & 0.5264 & 0.4621 & 0.5288 & 0.5135 & 0.5331 & 0.2711 & 0.3122 & -     & -     & - \\
        & maxF  & \best{0.7449} & 0.6963 & \second{0.7012} & 0.6650 & 0.6305 & 0.6265 & 0.5356 & 0.6097 & 0.6162 & 0.6274 & 0.5573 & 0.3810 & -     & -     & - \\
        & AUC   & \best{0.9516} & 0.9137 & \second{0.9284} & 0.9335 & 0.8937 & 0.8990 & 0.7956 & 0.8527 & 0.8602 & 0.8869 & 0.8656 & 0.6820 & -     & -     & - \\
        & MAE   & \best{0.0758} & 0.0963 & \second{0.0777} & 0.0978 & 0.1119 & 0.1113 & 0.1675 & 0.1409 & 0.1568 & 0.1210 & 0.3116 & 0.1761 & -     & -     & 0.1960 \\
        \hline
        \multirow{4}[1]{*}{ECSSD} & aveF  & \best{0.7589} & 0.7331 & \second{0.7369} & 0.6884 & 0.6417 & 0.6539 & 0.5535 & 0.6498 & 0.5927 & 0.6549 & 0.1853 & 0.3785 & 0.7270 & -     & - \\
        & maxF  & \best{0.8095} & \second{0.7928} & 0.7756 & 0.7390 & 0.6872 & 0.6987 & 0.6240 & 0.7012 & 0.6983 & 0.7037 & 0.6166 & 0.4493 & -     & -     & - \\
        & AUC   & \best{0.9009} & \second{0.8968} & 0.8736 & 0.8770 & 0.8398 & 0.8555 & 0.7655 & 0.8344 & 0.8293 & 0.8495 & 0.7986 & 0.6443 & -     & -     & - \\
        & MAE   & \best{0.1601} & \second{0.1602} & 0.1623 & 0.1841 & 0.2051 & 0.2127 & 0.2382 & 0.2040 & 0.2064 & 0.2055 & 0.3384 & 0.2859 & 0.1910 & -     & 0.1830 \\
        \hline
        \multirow{4}[2]{*}{ASD} & aveF  & 0.8932 & 0.8881 & \best{0.9067} & 0.8803 & 0.8787 & 0.8589 & 0.8196 & 0.8917 & 0.8516 & 0.8910 & 0.4141 & 0.6688 & -     & -     & - \\
        & maxF  & \best{0.9380} & \second{0.9345} & 0.9301 & 0.9204 & 0.9142 & 0.8935 & 0.8446 & 0.9114 & 0.8953 & 0.9142 & 0.8141 & 0.7100 & -     & -     & - \\
        & AUC   & \best{0.9913} & \second{0.9904} & 0.9871 & 0.9895 & 0.9786 & 0.9818 & 0.9483 & 0.9736 & 0.9654 & 0.9768 & 0.9512 & 0.8619 & -     & 0.9828 & - \\
        & MAE   & \best{0.0273} & 0.0292 & \second{0.0281} & 0.0353 & 0.0400 & 0.0601 & 0.0788 & 0.0488 & 0.0524 & 0.0431 & 0.1851 & 0.1466 & -     & -     & 0.0860 \\
        \hline
        \multirow{4}[1]{*}{PASCAL-S} & aveF  & \best{0.7310} & \second{0.7300} & 0.7177 & 0.6487 & 0.6415 & 0.6172 & 0.4151 & 0.6156 & 0.5504 & 0.6304 & 0.1713 & 0.3707 & 0.6690 & -     & - \\
        & maxF  & \best{0.8182} & \second{0.8051} & 0.7677 & 0.7307 & 0.7049 & 0.6782 & 0.6384 & 0.6893 & 0.6926 & 0.7097 & 0.6639 & 0.4837 & -     & -     & - \\
        & AUC   & \best{0.9287} & \second{0.9185} & 0.8742 & 0.8810 & 0.8482 & 0.8403 & 0.7995 & 0.8143 & 0.8267 & 0.8514 & 0.8184 & 0.6181 & -     & 0.8682 & - \\
        & MAE   & \best{0.1695} & 0.1818 & 0.1888 & 0.2351 & 0.2307 & 0.2565 & 0.2655 & 0.2464 & 0.2376 & 0.2385 & 0.3184 & 0.3297 & \second{0.1700} & -     & 0.2250 \\
        \hline
        \multirow{4}[0]{*}{SED2} & aveF  & \best{0.7778} & 0.7630 & 0.7766 & 0.7479 & \second{0.7776} & 0.7301 & 0.6648 & 0.7334 & 0.7024 & 0.7293 & 0.4083 & 0.6324 & -     & -     & - \\
        & maxF  & \best{0.8634} & \second{0.8448} & 0.8141 & 0.8386 & 0.8296 & 0.7890 & 0.7337 & 0.7670 & 0.7837 & 0.7710 & 0.7423 & 0.7104 & -     & -     & - \\
        & AUC   & \best{0.9557} & \second{0.9447} & 0.9036 & 0.9427 & 0.8814 & 0.9052 & 0.8354 & 0.8474 & 0.8448 & 0.8675 & 0.8666 & 0.7940 & -     & 0.9363 & - \\
        & MAE   & \best{0.1074} & \second{0.1142} & 0.1223 & 0.1228 & 0.1333 & 0.1452 & 0.1800 & 0.1567 & 0.1279 & 0.1512 & 0.2094 & 0.1901 & -     & -     & - \\

        \hline
        \multirow{4}[1]{*}{SOD} & aveF  & \best{0.6978} & \second{0.6910} & 0.6786 & 0.6023 & 0.6012 & 0.5980 & 0.4632 & 0.5710 & 0.5117 & 0.5902 & 0.1540 & 0.3535 & 0.6300 & -     & - \\
        & maxF  & \best{0.7807} & \second{0.7659} & 0.7262 & 0.6768 & 0.6530 & 0.6543 & 0.5551 & 0.6421 & 0.6456 & 0.6572 & 0.6242 & 0.4408 & -     & -     & - \\
        & AUC   & \best{0.9233} & \second{0.9115} & 0.8612 & 0.8624 & 0.8203 & 0.8409 & 0.7178 & 0.7950 & 0.8108 & 0.8382 & 0.8080 & 0.6004 & -     & 0.8477 & - \\
        & MAE   & \best{0.1503} & \second{0.1619} & 0.1750 & 0.2163 & 0.2136 & 0.2190 & 0.2523 & 0.2303 & 0.2297 & 0.2146 & 0.3610 & 0.2835 & 0.2050 & -     & - \\
        \hline
        \multirow{4}[2]{*}{THUR} & aveF  & \best{0.6254} & 0.6116 & \second{0.6189} & 0.5798 & 0.5266 & 0.5422 & 0.4732 & 0.5396 & 0.5091 & 0.5543 & 0.3175 & 0.3389 & -     & -     & - \\
        & maxF  & \best{0.7276} & \second{0.7173} & 0.6858 & 0.6702 & 0.5962 & 0.6107 & 0.5331 & 0.5972 & 0.5852 & 0.6096 & 0.5537 & 0.3861 & -     & -     & - \\
        & AUC   & \best{0.9567} & \second{0.9541} & 0.9239 & 0.9379 & 0.8865 & 0.9020 & 0.8027 & 0.8556 & 0.8535 & 0.8950 & 0.8655 & 0.6837 & -     & -     & - \\
        & MAE   & \best{0.0854} & \second{0.0897} & 0.0924 & 0.1050 & 0.1239 & 0.1190 & 0.1691 & 0.1421 & 0.1582 & 0.1254 & 0.2737 & 0.1775 & -     & -     & - \\
        \hline
        \multirow{4}[2]{*}{THUS} & aveF  & \best{0.8630} & -     & \multicolumn{1}{c|}{-} & \second{0.8514} & 0.8357 & 0.8095 & 0.7648 & 0.8322 & 0.8006 & 0.8338 & 0.4185 & 0.5933 & -     & -     & - \\
        & maxF  & \best{0.8994} & -     & \multicolumn{1}{c|}{-} & \second{0.8807} & 0.8559 & 0.8346 & 0.7938 & 0.8469 & 0.8449 & 0.8476 & 0.7895 & 0.6353 & -     & -     & - \\
        & AUC   & \best{0.9810} & -     & \multicolumn{1}{c|}{-} & \second{0.9776} & 0.9547 & 0.9588 & 0.9116 & 0.9435 & 0.9325 & 0.9507 & 0.9299 & 0.7896 & -     & -     & - \\
        & MAE   & \best{0.0628} & -     & \multicolumn{1}{c|}{-} & \second{0.0684} & 0.0845 & 0.1052 & 0.1185 & 0.0963 & 0.0920 & 0.0932 & 0.1831 & 0.1925 & -     & -     & - \\
        \hline
    \end{tabular}%
}
\caption{Comparison of average F-measure using adaptive threshold (aveF), maximum F-measure of average precision recall curve (maxF), AUC scores and MAE scores (smaller better).
    Our approach (OurLO) achieves the best performance in all these metrics. The results from the last three columns are directly quoted from their original papers.
    Since using THUS for training, the saliency results of OurBR are null.
}
\label{tab:comp}%
\end{table*}
\subsubsection{Quantitative performance comparison}
\begin{figure*}[htbp]
\centering 
\includegraphics[width=1\textwidth]{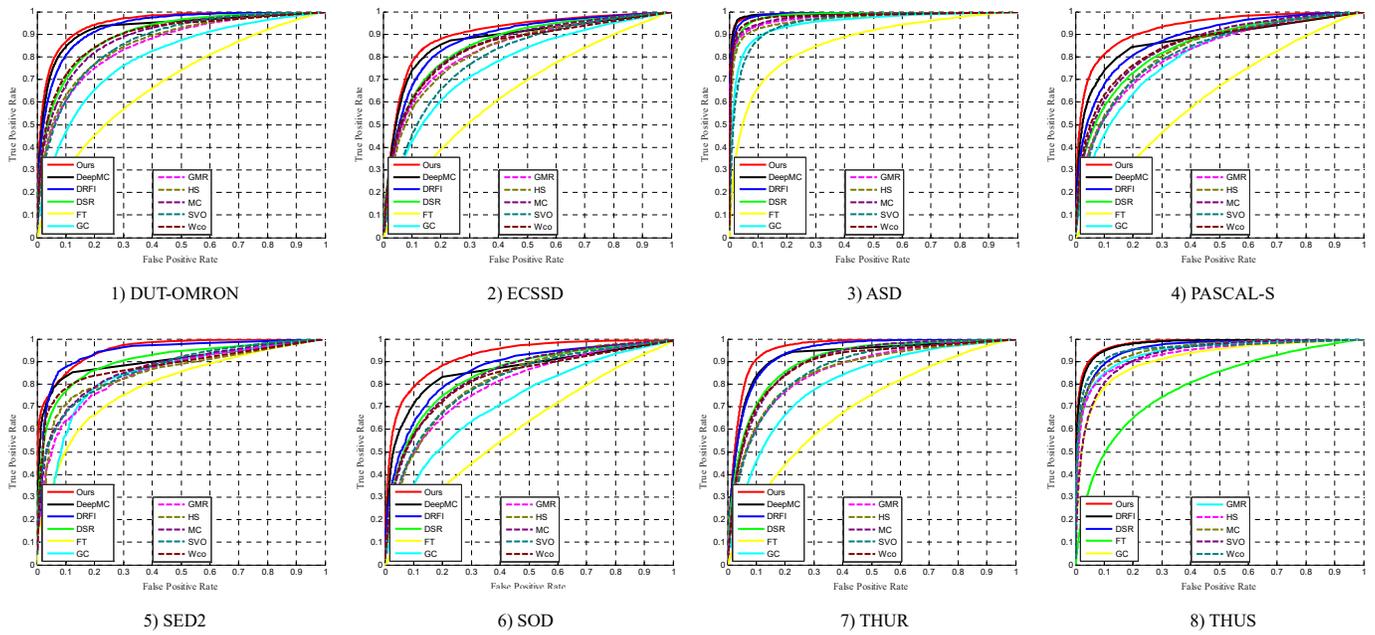}
\centering
\caption{ROC curves on the eight benchmark datasets. Clearly, our approach performs best in most cases.}
\label{fig:ROCcurve}
\vspace{-1.0\baselineskip}
\end{figure*}

Figure~\ref{fig:prCurve} shows
the corresponding PR curve performance of all the competing approaches
on the eight benchmark datasets.
%
We show the ROC curves in Figure~\ref{fig:ROCcurve}. From
Figure~\ref{fig:ROCcurve},
we observe that the proposed approach achieves a better performance
than the other ones in most cases.

More specifically, Table~\ref{tab:comp}
reports their quantitative saliency detection performance w.r.t.
the four evaluation metrics (i.e., aveF, maxF, AUC, and MAE) on
the eight benchmark datasets.
From Figure~\ref{fig:prCurve} and
Table~\ref{tab:comp}, it is clearly seen that our approach performs
favorably against the state-of-the-art methods in most cases.
%

\subsubsection{Qualitative performance comparison}
\begin{figure*}[htbp]
\centering 
\includegraphics[width=0.95\textwidth]{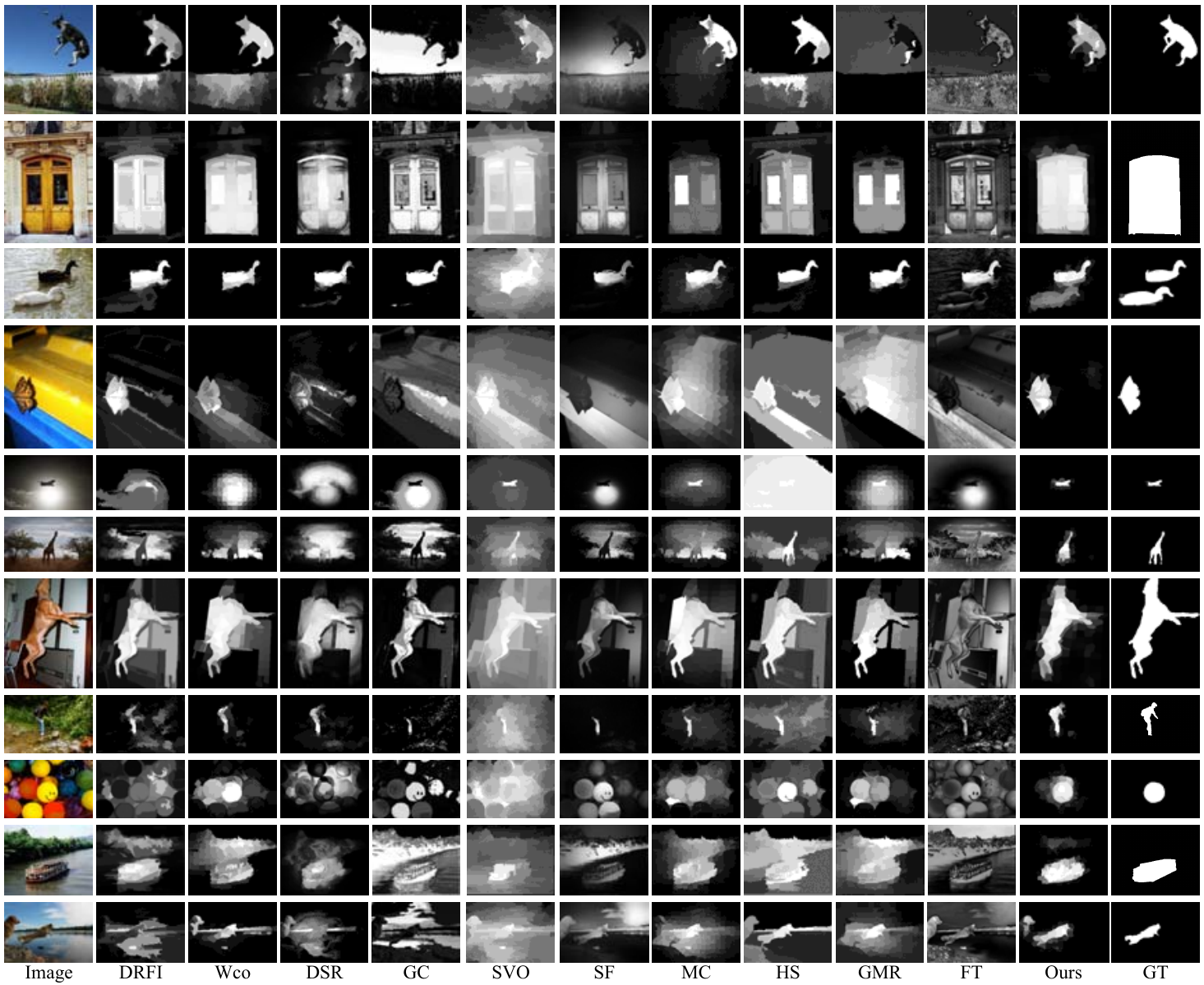}
\centering
\vspace{-0.5\baselineskip}
   \caption{
   Qualitative comparison of different approaches on several challenging samples with ground truth (GT). Clearly, our approach  obtains more visually feasible saliency detection results than the comparison approaches.
   }
\label{fig:QualititiveShow}
\vspace{-1.0\baselineskip}
\end{figure*}

For an intuitive illustration, we provide the saliency detection
results of our approach over several challenging sample images against
the other state-of-the-art approaches.
Figure~\ref{fig:QualititiveShow} shows that
our approach is able to obtain favorable saliency detection results
than the other methods. For example, as shown in the third and eighth rows,
our approach still works well in the
cases of background clutter and low foreground-background contrast.
%
%
%
%
%
%
%
%
%
%
%
%
%

\subsection{Analysis of Proposed Approach}

\subsubsection{Training strategies of FCNN}
We present the details on how to generate the training
data for FCNN in our saliency detection algorithm.
Here, we take three different approaches to generate training data.
One is the leave-one-out strategy, the second is the baseline-reference strategy, and the third is the small-training-set stratergy, which are respectively
referred to as OurLO, OurBR and OurST.
Specifically, as for OurLO,
when we test the performance
for a given dataset, the saliency training data for this dataset
are derived from the images (including the annotated saliency maps)
from the other datasets.
As for OurBR, we just select the largest and most
representative dataset (i.e., THUS) as the baseline-reference training data,
and the other 7 datasets are used for testing.
As for OurST, evaluate the quantitative performance of the proposed approach using a relatively small training set,
which is exactly the same as that of the DRFI approach~\cite{DRFI}. Namely, the training set is generated
by selecting $2500$ images from the MSRA5000~\cite{MSRA} (also called MSRA-B) dataset.
We note that some datasets are partly overlapped with the others, e.g., ASD and THUS, ECSSD and SOD.
Therefore, we make sure that the test data are disjoint
from the training data (by removing the overlapping images).

\begin{figure}[h]
    \centering
    \resizebox{0.95\linewidth}{!}{
        \includegraphics[width=1\textwidth]{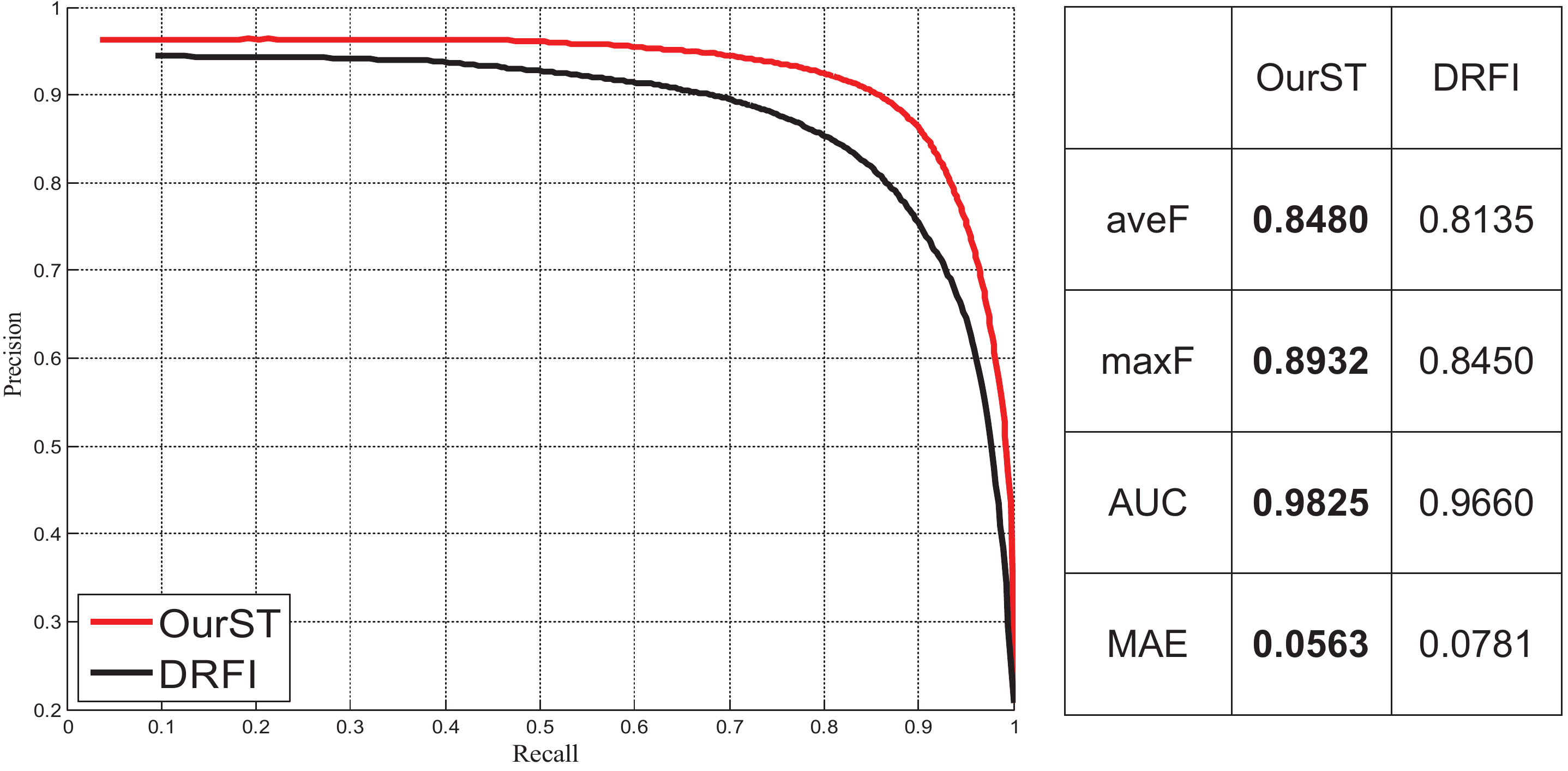}
    }
    \caption{Comparisons of OurST and DRFI methods on MSRA5000 dataset, including precision-recall curves and the results of four evaluation metrics.}
    \label{fig:msra5000}
    \vspace{-0.5\baselineskip}
\end{figure}

The third and fourth columns of Table~\ref{tab:comp} show the corresponding
quantitative saliency detection results of OurLO and OurBR, respectively.
Clearly, such two strategies achieve comparable saliency detection
results.
That is, both approaches perform similarly.
The performance of OurLO is slightly better than that of OurBR as
more training data are used.
Table~\ref{tab:smalldata} shows the quantitative results of OurLO, OurBR, OurST and DRFI on datasets which are not overlapped with the training set. We also report the results of OurST and DRFI on the MSRA5000 dataset (using the same testing list as that of DRFI with $2000$ images), as shown in Figure~\ref{fig:msra5000}.
From the above results, we observe that OurST still obtains comparable results with OurLO and OurBR using more training data, and also performs better than DRFI in most cases.
For consistency, our approach always refers to OurLO in the following.
%
%
%
\begin{table}[htbp]
  \centering
    \resizebox{0.95\linewidth}{!}{
    \begin{tabular}{|c|c|c|c|c|c|c|c|c|}
    \hline
        \multirow{2}[0]{*}{} & \multicolumn{4}{c|}{DUT-OMRON}  & \multicolumn{4}{c|}{ECSSD} \\
    \hline
          & aveF  & maxF  & AUC   & MAE   & aveF  & maxF  & AUC   & MAE   \\\hline
    OurLO & \textbf{0.6045 } & \textbf{0.7449 } & \textbf{0.9516 } & \textbf{0.0758 } & \textbf{0.7589 } & \textbf{0.8095 } & \textbf{0.9009 } & \textbf{0.1601 } \\
    OurBR & 0.5947  & 0.6963  & 0.9137  & 0.0963  & 0.7331  & 0.7928  & 0.8968  & 0.1602  \\
    OurST & 0.5824  & 0.7290  & 0.9454  & 0.0828  & 0.7365  & 0.7987  & 0.8975  & 0.1602  \\
    DRFI  & 0.5519  & 0.6650  & 0.9335  & 0.0978  & 0.6884  & 0.7390  & 0.8770  & 0.1841  \\
    \hline
    \end{tabular}%
    }
  \caption{Evaluations of different approaches including OurLO, OurBR, OurST and DRFI. The method OurST uses a small training set (i.e., $2500$ images from the MSRA5000 dataset).
    We observe that OurST still obtains comparable results with OurLO and OurBR using more training data,
    and also performs better than DRFI in most cases.}
  \label{tab:smalldata}%
\vspace{-0.5\baselineskip}
\end{table}%
\subsubsection{Evaluation on multi-task learning}\label{sec:single-task}
%
%
%
\begin{table}[t]\footnotesize
    \centering
    \resizebox{0.925\linewidth}{!}{
        \begin{tabular}{|c|c|c|c|c|c|}
            \hline
            & method & aveF  & maxF  & AUC   & MAE \\
            \hline
            \multirow{2}[2]{*}{DUT-OMRON} & Multi-task & \textbf{0.5947 } & \textbf{0.6963 } & 0.9137  & \textbf{0.0963 } \\
            & Single-task & 0.5171  & 0.6625  & \textbf{0.9147 } & 0.1155  \\
            \hline
            \multirow{2}[2]{*}{ECSSD} & Multi-task & \textbf{0.7331 } & \textbf{0.7928 } & \textbf{0.8968 } & \textbf{0.1602 } \\
            & Single-task & 0.7214  & 0.7852  & 0.8914  & 0.1651  \\
            \hline
            \multirow{2}[2]{*}{ASD} & Multi-task & \textbf{0.8881 } & \textbf{0.9345 } & \textbf{0.9904 } & \textbf{0.0292 } \\
            & Single-task & 0.8695  & 0.9201  & 0.9877  & 0.0367  \\
            \hline
            \multirow{2}[2]{*}{PASCAL-S} & Multi-task & \textbf{0.7300 } & \textbf{0.8051 } & \textbf{0.9185 } & 0.1818  \\
            & Single-task & 0.7120  & 0.7995  & 0.9178  & \textbf{0.1802 } \\
            \hline
            \multirow{2}[2]{*}{SED2} & Multi-task & \textbf{0.7630 } & \textbf{0.8448 } & \textbf{0.9447 } & \textbf{0.1142 } \\
            & Single-task & 0.7304  & 0.8272  & 0.9351  & 0.1240  \\
            \hline
            \multirow{2}[2]{*}{SOD} & Multi-task & \textbf{0.6910 } & \textbf{0.7659 } & 0.9115  & \textbf{0.1619 } \\
            & Single-task & 0.6726  & 0.7607  & \textbf{0.9127 } & 0.1623  \\
            \hline
            \multirow{2}[2]{*}{THUR} & Multi-task & \textbf{0.6116 } & \textbf{0.7173 } & \textbf{0.9541 } & \textbf{0.0897 } \\
            & Single-task & 0.5998  & 0.7163  & 0.9528  & 0.0898  \\
            \hline
            \hline
            \multirow{2}[2]{*}{\textbf{Average}} & Multi-task & \textbf{0.7159 } & \textbf{0.7938 } & \textbf{0.9328 } & \textbf{0.1190 } \\
            & Single-task & 0.6890  & 0.7816  & 0.9303  & 0.1248  \\
            \hline
        \end{tabular}%
    }
    \caption{
        Comparison of the proposed approach with (Multi-task, same as OurBR) and without multi-task learning on the
        seven datasets (except THUS). Clearly, our approach with multi-task learning achieves a better performance in most cases.}
    \label{tab:singleTask}%
\vspace{-1.5\baselineskip}
\end{table}%
We evaluate the performance differences of the proposed approach with and without
multi-task learning. Namely, the proposed approach without multi-task learning
is associated with that of only using the saliency-related network, resulting in
the single-task version.
By taking the same baseline-reference strategy in the paper,
we directly train
the model on the THUS dataset (excluding the overlapped part).
Table~\ref{tab:singleTask} shows the comparison results on the seven datasets (except THUS).
Clearly, multi-task learning helps to learn a better model for saliency detection in most cases, because performing semantic segmentation task improves the performance of saliency detection in object perception.

Semantic segmentation methods~\cite{FCN,crf2015ICLR,zheng2015crf} aim to segment pre-defined classes of objects in an image, no matter whether they are salient or not. For this reason, a state-of-the-art segmentation model~\cite{zheng2015crf} cannot be used for salient object detection, demonstrated by the experimental results in Table~\ref{tab:crfasrnn_comp}. For the segmentation part in our method, the saliency information can help detect and segment the salient objects to a certain extent in an image. Since the saliency region sizes are usually much smaller than those of the non-salient regions, the supervised learning information induced by salient object detection has a relatively small influence on the semantic segmentation task.
The experimental results generated from the proposed network demonstrate the above observations that the segmentations change trivially with and without joint training.
%
%
%
\begin{table}[htbp]
    \centering
        \begin{tabular}{|c|c|c|c|c|}
            \hline
            method & aveF  & maxF  & AUC   & MAE \\
            \hline
            Ours  & \textbf{0.8932} & \textbf{0.9380} & \textbf{0.9913} & \textbf{0.0273} \\
            CRFasRNN & 0.5475 & 0.6317 & 0.7883 & 0.1527  \\
            \hline
        \end{tabular}%
    \caption{Quantitative comparison results between the segmentation method~\cite{zheng2015crf} and ours over the ASD dataset.}
    \label{tab:crfasrnn_comp}%
\vspace{-0.8\baselineskip}
\end{table}%
\subsubsection{Evaluation on regression-based propagation}\label{sec:regression_crf}
%
%
%
%
\begin{table}[htbp]\scriptsize
    \centering
    \resizebox{0.95\linewidth}{!}{
        \begin{tabular}{|c|c|c|c|c|c|}\hline
            & method & aveF  & maxF  & AUC   & MAE \\
            \hline
            \multirow{2}[0]{1cm}{ DUT-OMRON} & with Pro & \textbf{0.6045 } & \textbf{0.7449 } & \textbf{0.9516 } & \textbf{0.0758 } \\
            & w/o Pro & 0.5970  & 0.7155  & 0.9502  & 0.0833  \\
            \hline
            \multirow{2}[0]{*}{ECSSD} & with Pro & \textbf{0.7589 } & \textbf{0.8095 } & \textbf{0.9009 } & 0.1601  \\
            & w/o Pro & 0.7245  & 0.8031  & 0.8905  & \textbf{0.1463 } \\
            \hline
            \multirow{2}[0]{*}{ASD} & with Pro & \textbf{0.8932 } & \textbf{0.9380 } & \textbf{0.9913 } & \textbf{0.0273 } \\
            & w/o Pro & 0.8500  & 0.9055  & 0.9883  & 0.0407  \\
            \hline
            \multirow{2}[0]{*}{PASCAL-S} & with Pro & \textbf{0.7310 } & \textbf{0.8182}  & \textbf{0.9287 } & 0.1695  \\
            & w/o Pro & 0.6266  & 0.8087 & 0.9191  & \textbf{0.1499 } \\
            \hline
            \multirow{2}[0]{*}{SED2}  & with Pro & \textbf{0.7778 } & \textbf{0.8634 } & 0.9557  & 0.1074  \\
            & w/o Pro & 0.7502  & 0.8431  & \textbf{0.9630 } & \textbf{0.0965 } \\
            \hline
            \multirow{2}[0]{*}{SOD}   & with Pro & \textbf{0.6978 } & \textbf{0.7807 } & 0.9233  & 0.1503  \\
            & w/o Pro & 0.6711  & 0.7774  & \textbf{0.9239 } & \textbf{0.1280 } \\
            \hline
            \multirow{2}[0]{*}{THUR}  & with Pro & \textbf{0.6254 } & 0.7276  & \textbf{0.9567 } & \textbf{0.0854 } \\
            & w/o Pro & 0.6209  & \textbf{0.7309 } & 0.9561  & 0.0908  \\
            \hline
            \multirow{2}[0]{*}{THUS}  & with Pro & \textbf{0.8630 } & \textbf{0.8994 } & \textbf{0.9810 } & \textbf{0.0628 } \\
            & w/o Pro & 0.7511  & 0.8336  & 0.9630  & 0.0973  \\
            \hline\hline
            \multirow{2}[0]{*}{Average} & with Pro & \textbf{0.7439 } & \textbf{0.8227 } & \textbf{0.9487 } & 0.1048  \\
            & w/o Pro & 0.6989  & 0.8022  & 0.9443  & \textbf{0.1041 } \\
            \hline
        \end{tabular}
    }
    \caption{Comparison of our approach with (with Pro) and without (w/o Pro) non-linear regression-based propagation.
        In most cases, our approach with propagation achieves a better performance.}
\vspace{-1.5\baselineskip}
\label{tab:propagation}%
\end{table}%
We quantitatively compare our approach with and without nonlinear
regression-based propagation on the eight benchmark datasets.
Table~\ref{tab:propagation} shows that
our approach with propagation will lead to a better saliency detection performance in most cases compared with that without propagation.
%
This can be attributed to the fact that nonlinear
regression-based propagation can capture more topological information
(among super-pixels) that is helpful to the boundary preservation of salient objects.

Since Conditional Random Fields (CRF) have been broadly used in recent semantic segmentation  methods~\cite{crf2015ICLR,zheng2015crf}, we also conduct experiments to compare the proposed refinement process with the one based on CRF. For the CRF one, the probability maps of the foreground class are taken as the saliency results. For the proposed method, we solve a regression problem to propagate the given information over the image, thus obtain refined maps. Same metrics are used for evaluating the resulted maps. We use the publicly available code of ~\cite{crf2015ICLR} and ~\cite{krahenbuhl2011efficient} to implement the algorithm.
%
%
%
\begin{table}[htbp]
    \centering
    \resizebox{0.925\linewidth}{!}{
    \begin{tabular}{|c|c|c|c|c|c|}
        \hline
        & method & aveF  & maxF  & AUC   & MAE \\
        \hline
        \multirow{3}[2]{1cm}{DUT-
            OMRON} & OnlyDeep & 0.5970 & 0.7155 & 0.9502 & 0.0833 \\
        & Ours  & 0.6045 & 0.7449 & \textbf{0.9516} & 0.0758 \\
        & CRF   & \textbf{0.7022} & \textbf{0.7603} & 0.8289 & \textbf{0.0644} \\
        \hline
        \multirow{3}[2]{*}{ECSSD} & OnlyDeep & 0.7245 & 0.8031 & 0.8905 & 0.1463 \\
        & Ours  & 0.7589 & 0.8095 & \textbf{0.9009} & 0.1601 \\
        & CRF   & \textbf{0.7710} & \textbf{0.8240} & 0.8172 & \textbf{0.1390} \\
        \hline
        \multirow{3}[2]{*}{MSRA1000} & OnlyDeep & 0.8500 & 0.9055 & 0.9883 & 0.0407 \\
        & Ours  & 0.8932 & 0.9380 & \textbf{0.9913} & 0.0273 \\
        & CRF   & \textbf{0.9235} & \textbf{0.9422} & 0.9646 & \textbf{0.0245} \\
        \hline
        \multirow{3}[2]{*}{PASCAL-S} & OnlyDeep & 0.6266 & 0.8087 & 0.9191 & 0.1499 \\
        & Ours  & \textbf{0.7310} & \textbf{0.8182} & \textbf{0.9287} & 0.1695 \\
        & CRF   & 0.6679 & 0.8139 & 0.8354 & \textbf{0.1361} \\
        \hline
        \multirow{3}[2]{*}{SED2} & OnlyDeep & 0.7502 & 0.8431 & \textbf{0.9630} & 0.0965 \\
        & Ours  & 0.7778 & 0.8634 & 0.9557 & 0.1074 \\
        & CRF   & \textbf{0.8205} & \textbf{0.8768} & 0.8337 & \textbf{0.0949} \\
        \hline
        \multirow{3}[2]{*}{SOD} & OnlyDeep & 0.6711 & 0.7774 & \textbf{0.9239} & \textbf{0.1280} \\
        & Ours  & \textbf{0.6978} & \textbf{0.7807} & 0.9233 & 0.1503 \\
        & CRF   & 0.6459 & 0.7737 & 0.7635 & 0.1361 \\
        \hline
        \multirow{3}[2]{*}{THUR} & OnlyDeep & 0.6209 & 0.7309 & 0.9561 & 0.0908 \\
        & Ours  & 0.6254 & 0.7276 & \textbf{0.9567} & 0.0854 \\
        & CRF   & \textbf{0.6992} & \textbf{0.7472} & 0.8596 & \textbf{0.0780} \\
        \hline
        \multirow{3}[2]{*}{THUS} & OnlyDeep & 0.7511 & 0.8336 & 0.9630 & 0.0973 \\
        & Ours  & \textbf{0.8630} & \textbf{0.8994} & \textbf{0.9810} & \textbf{0.0628} \\
        & CRF   & 0.8207 & 0.8789 & 0.9041 & 0.0696 \\
        \hline
        \hline
        \multirow{3}[2]{*}{\textbf{Average}} & OnlyDeep & 0.6989 & 0.8022 & 0.9443 & 0.1041 \\
        & Ours  & 0.7439 & 0.8227 & \textbf{0.9487} & 0.1048 \\
        & CRF   & \textbf{0.7563} & \textbf{0.8271} & 0.8509 & \textbf{0.0928} \\
        \hline
    \end{tabular}%
    }
    \caption{
        Comparison between the proposed refinement method with the one based on CRF model on eight benchmark datasets.}
    \label{tab:crf}%
\vspace{-0.5\baselineskip}
\end{table}%
Table~\ref{tab:crf} shows that the performances of these two methods are comparable, while our method outperforms the CRF one on three datasets, including the largest dataset THUS. Both two methods can be used for refinement by solving an optimization problem given a graph on the image, and the proposed method is very simple and efficient as a closed-form solution exists. Furthermore, the gradient w.r.t the network output can be easily computed based on Eq.~\ref{eq:finalAlpha} for back-propagation, and then an end-to-end learning architecture with refinement, like~\cite{zheng2015crf}, is also possible for future research.

\subsubsection{Effect of $\beta$}
\begin{table}[htbp]\footnotesize
    \centering
    \resizebox{0.925\linewidth}{!}{
        \begin{tabular}{|c|c|c|c|c|c|}\hline
            & 0     & 0.1   & 0.2   & 0.3   & 0.5 \\
            \hline
            aveF  & 0.8689 & 0.8783 & 0.8932 & 0.9043 & \textbf{0.9143} \\
            maxF  & 0.9195 & 0.9352 & \textbf{0.9380} & 0.9368 & 0.9318 \\
            AUC   & 0.9880 & 0.9910 & \textbf{0.9913} & 0.9910 & 0.9895 \\
            MAE   & 0.0329 & 0.0284 & \textbf{0.0273} & 0.0304 & 0.0431 \\
            \hline
        \end{tabular}%
    }
    \caption{Evaluation of our saliency detection approach using
        different value of $\beta$.
        %
        %
        Clearly, the performance of our approach is relatively stable
        w.r.t. different choices of $\beta$, and we choose $\beta = 0.2$ in our experiments.
    }
\vspace{-1.5\baselineskip}
\label{tab:beta}%
\end{table}%
%
%
%
To analyze the relative effect of nonlinear saliency regression
using deep saliency learning and image boundary propagation, we
perform quantitative experiments (on the ASD dataset) to
evaluate the saliency detection performance w.r.t.
different configurations of the trade-off control factors $\beta$ such that
$\beta \in \{0.0,0.1,0.2,0.3,0.5\}$, as shown in Table~\ref{tab:beta}.
The results show that the setting of $\beta=0.2$ leads to
a better performance in most cases. Moreover, the performance of
our approach keeps relatively stable w.r.t. different choices of
$\beta$. When $\beta = 0$, only the saliency information from FCNN
is used. With the increase of $\beta$, our saliency performance
gradually improves until  $\beta = 0.2$.
After that, the performance keeps relatively stable.
Therefore, our deep saliency learning part plays a crucial
role in capturing the semantic object properties, while image boundary
propagation can result in a slight improvement
due to introducing more spatially structural information.
The trade-off control factors $\beta$ for coarse-grained
saliency fusion is set to $0.2$ in all the experiments.


\section{Conclusion}

In this paper, we propose a simple yet effective multi-task deep saliency approach for salient
object detection based on the fully convolutional neural network
with global input (whole raw images)
and global output (whole saliency maps).
The proposed saliency approach
models the
intrinsic semantic properties of salient objects in a totally data-driven
manner, and
performs collaborative feature learning
for the two correlated tasks (i.e., saliency detection
and semantic image segmentation), which generally leads to
the saliency performance improvement in object
perception.
Moreover, it
is capable of accomplishing the feature-sharing task
by using a sequence of fully convolutional layers,
resulting in
a significant reduction of feature redundancy.
In order to obtain more fine-grained saliency
detection results, we present a saliency
refinement method based on graph Laplacian regularized nonlinear
regression with a closed-form solution, which aims to propagate the saliency information over the spatially-adjacent
super-pixel graph for further saliency performance enhancement.
Experimental results on the eight benchmark datasets
demonstrate the proposed approach performs favorably in different
evaluation metrics against the state-of-the-art methods.

%
%
%





\bibliographystyle{IEEEtran}
\bibliography{saliency_tip}
%
\vspace{-1.5\baselineskip}
\begin{IEEEbiography}[{\includegraphics[width=1in,height=1.25in,clip,keepaspectratio]{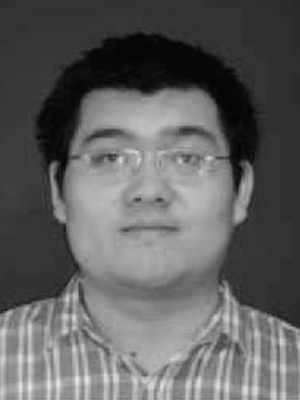}}]{Xi Li}
is currently a full professor at the Zhejiang University, China. Prior to that, he was a senior researcher at the University of Adelaide, Australia. From 2009 to 2010, he worked as a postdoctoral researcher at CNRS Telecomd ParisTech, France. In 2009, he got the doctoral degree from National Laboratory of Pattern Recognition, Chinsese Academy of Sciences, Beijing, China. His research interests include visual tracking, motion analysis, face recognition, web data mining, image and video retrieval.
\end{IEEEbiography}
\vspace{-1.5\baselineskip}
\begin{IEEEbiography}[{\includegraphics[width=1in,height=1.25in,clip,keepaspectratio]{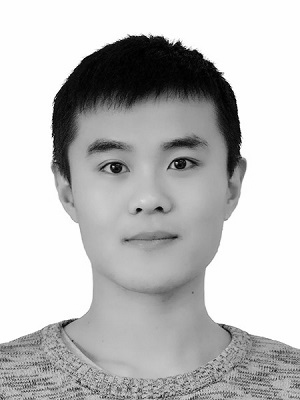}}]{Liming Zhao}
is currently a third year PhD student in College of Computer Science at Zhejiang University, Hangzhou, China. His advisors are Prof. Xi Li and Prof. Yueting Zhuang. Earlier, he received his bachelor's degree in Software Engineering from Shandong University, in 2013. His current research interests are primarily in computer vision and machine learning, especially deep learning, visual attention, object recognition, detection and segmentation.
\end{IEEEbiography}
\vspace{-1.5\baselineskip}
\begin{IEEEbiography}[{\includegraphics[width=1in,height=1.25in,clip,keepaspectratio]{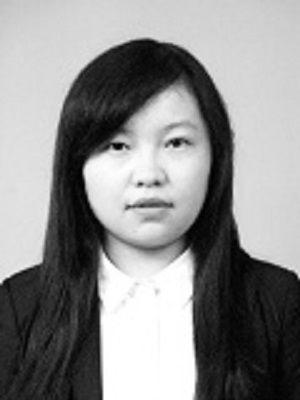}}]{Lina Wei}
received the B.S. degree from school of Information Science and Engineering, East China University of Science and Technology, Shanghai, China, in 2013. She is currently a third year PhD student in College of Computer Science at Zhengjiang University, China. Her advisors are Prof. Xi Li and Prof. Fei Wu. Her current research interests are primarily in image and video saliency detection.
\end{IEEEbiography}
\begin{IEEEbiography}[{\includegraphics[width=1in,height=1.25in,clip,keepaspectratio]{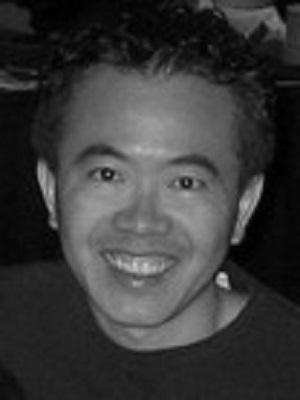}}]{Ming-Hsuan Yang}
received the PhD degree in computer science from the University of Illinois at Urbana-Champaign in 2000. He is an associate professor in electrical engineering and computer science at the University of California, Merced. Prior to joining UC Merced in 2008, he was a senior research scientist at the Honda Research Institute working on vision problems of humanoid robots. He served as an associate editor of the \textit{IEEE Transactions on Pattern Analysis and Machine Intelligence} from 2007 to 2011, and is an associate editor of the \textit{International Journal of Computer Vision, Image and Vision Computing}, and \textit{Journal of Artificial Intelligence Research}. He received the US National Science Foundation CAREER award in 2012, the Senate Award for Distinguished Early Career Research at UC Merced in 2011, and the Google Faculty Award in 2009. He is a senior member of the IEEE and the ACM.
\end{IEEEbiography}
\vspace{-1.5\baselineskip}
\begin{IEEEbiography}[{\includegraphics[width=1in,height=1.25in,clip,keepaspectratio]{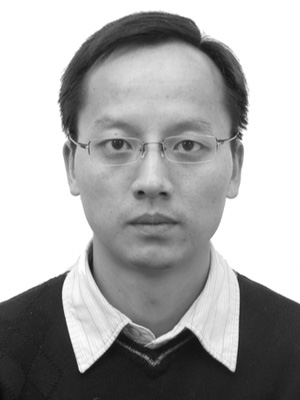}}]{Fei Wu}
	received the B.S. degree from Lanzhou University, Lanzhou, Gansu, China, the M.S. degree from Macao University, Taipa, Macau, and the Ph.D. degree from Zhejiang University, Hangzhou, China. He is currently a Full Professor with the College of Computer Science and Technology, Zhejiang University. He was a Visiting Scholar with Prof. B. Yu's Group, University of California, Berkeley, from 2009 to 2010. His current research interests include multimedia retrieval, sparse representation, and machine learning.
\end{IEEEbiography}
\vspace{-1.5\baselineskip}
\begin{IEEEbiography}[{\includegraphics[width=1in,height=1.25in,clip,keepaspectratio]{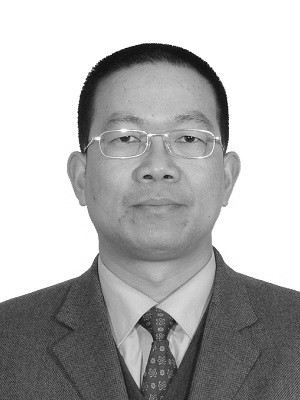}}]{Yueting Zhuang}
received the B.S., M.S., and Ph.D. degrees from Zhejiang University, Hangzhou, China, in 1986, 1989, and 1998, respectively. From 1997 to 1998, he was a visitor at the Department of Computer Science and Beckman Institute, University of Illinois at UrbanaChampaign, Champaign, IL, USA. Currently, he is a Professor and the Dean of the College of Computer Science, Zhejiang University. His current research interests include multimedia databases, artificial intelligence, and video-based animation.
\end{IEEEbiography}
\vspace{-1.5\baselineskip}
\begin{IEEEbiography}[{\includegraphics[width=1in,height=1.25in,clip,keepaspectratio]{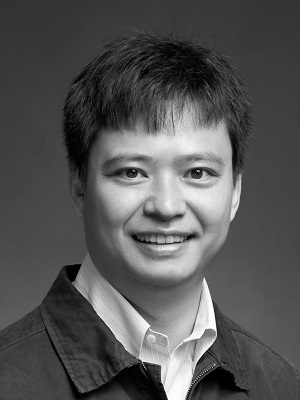}}]{Haibin Ling}
received the B.S. and M.S. degrees from Peking University, in 1997 and 2000, respectively, and the Ph.D. degree from the University of Maryland, College Park, in 2006. From 2000 to 2001, he was an Assistant Researcher with Microsoft Research Asia. From 2006 to 2007, he was a Post-Doctoral Scientist with the University of California at Los Angeles. He joined Siemens Corporate Research as a Research Scientist. Since fall 2008, he has been with Temple University, where he is currently an Associate Professor. His research interests include computer vision, medical image analysis, and human computer interaction. He received the Best Student Paper Award at the ACM Symposium on User Interface Software and Technology in 2003, and the NSF CAREER Award in 2014. He has served as an Area Chair of the Conference on Computer Vision and Pattern Recognition in 2014, and a Guest Co-Editor of Pattern Recognition.
\end{IEEEbiography}
\vspace{-1.5\baselineskip}
\begin{IEEEbiography}[{\includegraphics[width=1in,height=1.25in,clip,keepaspectratio]{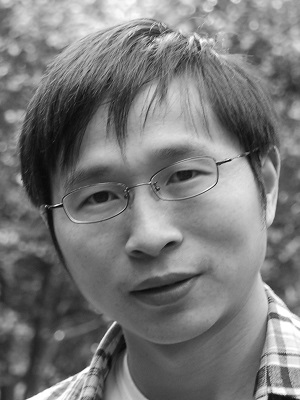}}]{Jingdong Wang}
received  the  B.S.  and  M.S. degrees in Automation from Tsinghua University, Beijing, China, in 2001 and 2004, respectively, and the Ph.D. degree in Computer Science from
the Hong Kong University of Science and Technology,  Hong  Kong,  in  2007.  He  is  currently a  researcher  at  the  Media  Computing  Group, Microsoft  Research  Asia.  His  areas  of  interest include computer vision, machine learning, and multimedia  search.  At  present,  he  is  mainly working  on  the  Big  Media  project,  including
large-scale indexing and clustering, and Web image search and mining. He is an editorial board member of Multimedia Tools and Applications.
\end{IEEEbiography}
\vfill
\end{document}